\PassOptionsToPackage{table}{xcolor}

\documentclass[nohyperref]{article}

\usepackage[accepted]{icml2023}

\usepackage{scalefnt,letltxmacro}
\LetLtxMacro{\oldtextsc}{\textsc}
\renewcommand{\textsc}[1]{\oldtextsc{\scalefont{1.10}#1}}
\usepackage[acronym,smallcaps,nowarn]{glossaries}
\glsdisablehyper
\makeglossaries
\usepackage{xspace}
\usepackage{float}
\usepackage{physics}
\usepackage{lipsum}

\usepackage{enumitem}

\usepackage{amssymb}
\usepackage{mathtools}
\usepackage{amsfonts}
\usepackage{amsmath}
\usepackage{amsthm}
\usepackage{booktabs}
\usepackage[]{microtype}

\usepackage{pifont}
\newcommand{\cmark}{\textcolor{green!60!black}{\ding{51}}\xspace}
\newcommand{\xmark}{\textcolor{red!60!black}{\ding{55}}\xspace}

\usepackage[vlined,linesnumbered,ruled]{algorithm2e}

\SetCommentSty{mycommfont}
\SetKwInput{KwInput}{Input} 
\SetKwInput{KwOutput}{Output}

\usepackage{xcolor}
\definecolor{mylightgray}{gray}{0.94}

\definecolor{color_vae}{HTML}{91058A}
\definecolor{color_cvae}{HTML}{DA70D6}
\definecolor{color_gppvae}{HTML}{32CD32}
\definecolor{color_gpvae}{HTML}{039796}
\definecolor{color_gp_vae}{HTML}{00FFFF}
\definecolor{color_sgp_vae}{HTML}{99CB32}
\definecolor{color_svgp_vae}{HTML}{19198c}
\definecolor{color_bae}{HTML}{AB6345}
\definecolor{color_gpbae}{HTML}{FF9B38}
\definecolor{color_bsgpae}{HTML}{DF2B4F}
\definecolor{color_blue_trajectory}{HTML}{DF2B4F}
\definecolor{color_blue_trajectory}{HTML}{3E9EFF}
\definecolor{color_orange_trajectory}{HTML}{DF5E33}

\usepackage[colorlinks,linktoc=all]{hyperref}
\usepackage[all]{hypcap}
\definecolor{frenchblue}{rgb}{0.01171875, 0.0078125, 0.4375}
\hypersetup{citecolor=frenchblue}
\hypersetup{linkcolor=frenchblue}
\hypersetup{urlcolor =frenchblue}

\usepackage[capitalize,nameinlink]{cleveref}
\crefname{section}{\S}{\S\S}
\Crefname{section}{\S}{\S\S}
\creflabelformat{equation}{#2\textup{#1}#3}

\makeatletter
\@ifundefined{c@rownum}{%
  \let\c@rownum\rownum
}{}
\@ifundefined{therownum}{%
  \def\therownum{\@arabic\rownum}%
}{}
\makeatother

\makeatletter
\newcommand*{\addFileDependency}[1]{%
	\typeout{(#1)}
	\@addtofilelist{#1}
	\IfFileExists{#1}{}{\typeout{No file #1.}}
}
\makeatother

\usepackage[font=small,labelfont=bf,tableposition=top]{caption}
\usepackage{tikz}
\usepackage{graphicx}
\usepackage[]{subcaption}   %

\usepackage{pgfplots}
\pgfplotsset{compat=1.6}
\usepgfplotslibrary{groupplots}
\tikzstyle{every picture}+=[font=\sffamily]
\tikzstyle{optimized} = [circle,fill=white,draw=black, dashed,inner sep=1pt, minimum size=20pt, font=\fontsize{10}{10}\selectfont, node distance=1]
\pgfkeys{/pgf/number format/.cd,1000 sep={}}
\pgfplotsset{
	tick label style = {font=\sffamily},
	every axis label/.append style={font=\sffamily},
	typeset ticklabels with strut,
}
\pgfplotsset{every axis/.append style={
			every x tick label/.append style={font=\fontsize{6pt}{6pt}\sffamily, yshift=.5ex,},
			every y tick label/.append style={font=\fontsize{6pt}{6pt}\sffamily, xshift=.5ex},
			every y label/.append style={xshift=10ex, font=\sffamily},
			every x label/.append style={yshift=3ex, font=\sffamily},
			every title/.append style={font=\sffamily}
		},
}
\pgfplotsset{
  xticklabel={$\mathsf{\pgfmathprintnumber{\tick}}$},
  yticklabel={$\mathsf{\pgfmathprintnumber{\tick}}$},
}
\pgfplotsset{every axis title/.append style={yshift=-1ex}}
\newlength\figureheight
\newlength\figurewidth
\usepgfplotslibrary{external}
\usepackage[colorinlistoftodos,
	textsize=scriptsize,
	linecolor=red!30,
	bordercolor=red!30,
	backgroundcolor=red!10]{todonotes}
\renewcommand{\todo}[2][]{\tikzexternaldisable\@todo[#1]{#2}\tikzexternalenable}

\usepackage[most]{tcolorbox}
{\begin{tcolorbox}[toprule=2mm,left=4pt,right=4pt,top=0pt,bottom=1pt,boxsep=2pt, before skip = 2ex, after skip =2ex]}{\end{tcolorbox}}
\tcbset{boxsep=4pt,left=2pt,right=2pt,top=-0pt,bottom=0pt}

\usepackage{url}

\usepackage{xr}
\usepackage{subcaption}

\usepackage{tikz}
\usepackage{graphicx}
\usepackage{xcolor}
\usepackage{amsmath}
\usepackage{amssymb}
\usepackage{bm}

\usepackage{adjustbox}
\usepackage{multirow}
\usepackage{mathtools}

\usepackage{xspace}

\newacronym{MAP}{map}{maximum-a-posteriori}
\newacronym{MLE}{mle}{maximum likelihood estimation}
\newacronym{MNLL}{mnll}{mean negative loglikelihood}
\newacronym{NLL}{nll}{negative loglikelihood}
\newacronym{LL}{ll}{log-likelihood}
\newacronym{RMSE}{rmse}{root mean squared error}
\newacronym{ECE}{ece}{expected calibration error}
\newacronym{FID}{fid}{Fr\'echet Inception Distance}
\newacronym{MSE}{mse}{mean squared error}

\newacronym{PCA}{pca}{principal component analysis}
\newacronym{AE}{ae}{autoencoder}
\newacronym{WAE}{wae}{Wasserstein Autoencoder}
\newacronym{VAE}{vae}{Variational Autoencoder}
\newacronym{BAE}{bae}{Bayesian autoencoder}
\newacronym{CDF}{cdf}{cumulative density function}
\newacronym{GAN}{gan}{Generative Adversarial Network}

\newacronym{MC}{mc}{Monte Carlo}
\newacronym{MCMC}{mcmc}{Markov chain Monte Carlo}
\newacronym{HMC}{hmc}{Hamiltonian Monte Carlo}
\newacronym{MH}{mh}{Metropolis-Hastings}
\newacronym{NUTS}{nuts}{no-u-turn sampler}
\newacronym{SGHMC}{sghmc}{stochastic gradient Hamiltonian Monte Carlo}

\newacronym{DGP}{dgp}{deep Gaussian process} %
\newacronym{GPLVM}{gplvm}{Gaussian process latent variable model}
\newacronym{DPMM}{dpmm}{Dirichlet Process Mixture Model}

\newacronym{VFE}{vfe}{variational free energy}

\newacronym[firstplural=Gaussian Processes]{GP}{gp}{Gaussian Process}

\newacronym{VI}{vi}{variational inference}

\newacronym{ELBO}{elbo}{evidence lower bound}
\newacronym{NELBO}{nelbo}{negative evidence lower bound}
\newacronym{ELL}{ell}{expected log likelihood}
\newacronym{KL}{kl}{Kullback-Leibler divergence}
\newacronym{AUC}{auc}{area under the curve}

\newacronym[firstplural=Bayesian neural networks]{BNN}{bnn}{Bayesian neural network}
\newacronym[firstplural=deep neural networks]{DNN}{dnn}{deep neural network}
\newacronym[]{CNN}{cnn}{convolutional neural network}
\newacronym{MLP}{mlp}{multilayer perceptron}
\newacronym{NN}{nn}{neural network}
\newacronym{RELU}{ReLU}{rectified linear unit}

\newacronym{NF}{nf}{normalizing flow}

\newacronym{RBF}{rbf}{radial basis function}
\newacronym{ARD}{ard}{automatic relevance determination}

\newacronym{RKHS}{rkhs}{reproducing kernel Hilbert space}

\newacronym{OT}{ot}{optimal transport}
\newacronym{WD}{wd}{Wasserstein distance}
\newacronym{SWD}{swd}{sliced-Wasserstein distance}
\newacronym{DSWD}{dswd}{distributional sliced-Wasserstein distance}
\newacronym{BSGPAE}{bsgpae}{Bayesian Sparse Gaussian Process Autoencoder}
\newacronym{GPBAE}{{gp}-{bae}}{Gaussian Process Bayesian Autoencoder}
\newacronym{CVAE}{cvae}{Conditional Variational Autoencoder}
\newacronym{SGPBAE}{{sgp}-{bae}}{Sparse Gaussian Process Bayesian Autoencoder}

\newcommand{\name}[1]{{\textsc{#1}}\xspace}

\newcommand{\mnist}{\name{mnist}}

\newcommand{\gpvae}{\textsc{gp}-\textsc{vae}\xspace}
\newcommand{\svgpvae}{\textsc{svgp}-\textsc{vae}\xspace}
\newcommand{\deepsvigp}{\textsc{deep}-\textsc{svigp}\xspace}

\newcommand{\deepsgpbae}{\textsc{dsgp}-\textsc{bae}\xspace}
\newcommand{\gppvae}{\textsc{gppvae}\xspace}
\newcommand{\sgpvae}{\textsc{sgp}-\textsc{vae}\xspace}
\newcommand{\igp}{\textsc{igp}\xspace}
\newcommand{\gpar}{\textsc{gpar}\xspace}
\newcommand{\nll}{\textsc{nll}\xspace}
\newcommand{\mae}{\textsc{mae}\xspace}
\newcommand{\smse}{\textsc{smse}\xspace}
\newcommand{\mathbold}[1]{{\boldsymbol{\mathbf{#1}}}}

\newcommand{\g}{\,|\,}
\makeatletter
\newcommand{\given}{\,\middle|\,}
\makeatother

\newcommand{\nestedmathbold}[1]{{\mathbold{#1}}}

\newcommand{\mbf}{\nestedmathbold{f}}

\newcommand{\mbr}{\nestedmathbold{r}}
\newcommand{\mbs}{\nestedmathbold{s}}

\newcommand{\mbu}{\nestedmathbold{u}}
\newcommand{\mbv}{\nestedmathbold{v}}

\newcommand{\mbx}{\nestedmathbold{x}}
\newcommand{\mby}{\nestedmathbold{y}}
\newcommand{\mbz}{\nestedmathbold{z}}

\newcommand{\mbB}{\nestedmathbold{B}}
\newcommand{\mbC}{\nestedmathbold{C}}

\newcommand{\mbI}{\nestedmathbold{I}}

\newcommand{\mbK}{\nestedmathbold{K}}

\newcommand{\mbM}{\nestedmathbold{M}}

\newcommand{\mbS}{\nestedmathbold{S}}

\newcommand{\mbX}{\nestedmathbold{X}}
\newcommand{\mbY}{\nestedmathbold{Y}}
\newcommand{\mbZ}{\nestedmathbold{Z}}

\newcommand{\mbphi}{\nestedmathbold{\phi}}

\newcommand{\mbtheta}{\nestedmathbold{\theta}}

\newcommand{\mbvarphi}{\nestedmathbold{\varphi}}

\newcommand{\mbPsi}{\nestedmathbold{\Psi}}

\DeclarePairedDelimiterX{\infdivx}[2]{[}{]}{%
  #1\;\delimsize\|\;#2%
}

\newcommand{\cL}{\mathcal{L}}
\newcommand{\cN}{\mathcal{N}}

\newcommand{\cI}{\mathcal{I}}

\newcommand{\E}{\mathbb{E}}

\newcommand{\bbR}{\mathbb{R}}

\newcommand{\Kxx}{\mbK_{\mbx \mbx}}
\newcommand{\Kss}{\mbK_{\mbs \mbs}}
\newcommand{\Kxs}{\mbK_{\mbx \mbs}}
\newcommand{\Ksx}{\mbK_{\mbs \mbx}}
\newcommand{\Kssinv}{\mbK_{\mbs \mbs}^{-1}}

\newcommand{\diag}{\textrm{diag}}

\theoremstyle{plain}

\theoremstyle{definition}

\theoremstyle{remark}

\icmltitlerunning{{Fully Bayesian Autoencoders with Latent Sparse Gaussian Processes}\hfill\thepage}

\begin{document}

\twocolumn[
\icmltitle{Fully Bayesian Autoencoders with Latent Sparse Gaussian Processes}

\icmlsetsymbol{equal}{*}

\begin{icmlauthorlist}
\icmlauthor{Ba-Hien Tran}{eurecom}
\icmlauthor{Babak Shahbaba}{uci}
\icmlauthor{Stephan Mandt}{uci}
\icmlauthor{Maurizio Filippone}{eurecom}
\end{icmlauthorlist}

\icmlaffiliation{eurecom}{Department of Data Science, EURECOM, France}
\icmlaffiliation{uci}{Departments of Statistics and Computer Science, University of California, Irvine, USA}

\icmlcorrespondingauthor{Ba-Hien Tran}{ba-hien.tran@eurecom.fr}
\icmlkeywords{Machine Learning, ICML}

\vskip 0.3in
]

\printAffiliationsAndNotice{}  %

\begin{abstract}
    Autoencoders and their variants are among the most widely used models in representation learning and generative modeling.
    However, autoencoder-based models usually assume that the learned representations are i.i.d. and fail to capture the correlations between the data samples. 
    To address this issue, we propose a novel Sparse Gaussian Process Bayesian Autoencoder (SGP-BAE) model in which we impose fully Bayesian sparse Gaussian Process priors on the latent space of a Bayesian Autoencoder.
    We perform posterior estimation for this model via stochastic gradient Hamiltonian Monte Carlo.
    We evaluate our approach qualitatively and quantitatively on a wide range of representation learning and generative modeling tasks and show that our approach consistently outperforms multiple alternatives relying on Variational Autoencoders.
\end{abstract}

\begin{table*}[t]
    \caption{A summary of related methods. Here, $\mbtheta, \mbu, \mbS$ refer to \gls{GP} hyper-parameters, inducing variables and inducing inputs, respectively.
        $N$ and $B$ are the number of data points and the mini-batch size, whereas $M, H \ll N$ are the number of inducing points and the  low-rank matrix factor, respectively.
        The colors denoting the methods shall be used consistently throughout the paper.
        } \label{table:comparison}
    \centering
    \vspace{-1ex}
    \resizebox{0.99\textwidth}{!}{%
        \begin{tabular}{lcccccccr}
            \toprule
            Model                                                                                                                                                                                                                & \begin{tabular}[c]{@{}c@{}} Scalable \\ (Minibatching) \end{tabular} & \begin{tabular}[c]{@{}c@{}} Non i.i.d. \\ data \end{tabular} & \begin{tabular}[c]{@{}c@{}} Free-form \\  posterior  \end{tabular} & \begin{tabular}[c]{@{}c@{}} \textsc{gp} \\  complexity  \end{tabular}  & \begin{tabular}[c]{@{}c@{}} Arbitrary  kernel \\ \& data type \end{tabular} & \begin{tabular}[c]{@{}c@{}} Learnable \\ \textsc{gp}  \end{tabular} & \begin{tabular}[c]{@{}c@{}} Inference \\ $\mbtheta, \mbu, \mbS$  \end{tabular} & Reference          \\
            \midrule
            \midrule
            \tikzexternaldisable ({\protect\tikz[baseline=-.65ex]\protect\draw[thick, color=color_vae, fill=color_vae, mark=*, mark size=2pt, line width=1.25pt] plot[] (-.0, 0)--(-0,0);})  \textsc{vae}                         & \cmark       & \xmark                    & \xmark                    & -                          & -                         & -                         & -                         & \citet{Kingma14}   \\
            \tikzexternalenable
            \tikzexternaldisable ({\protect\tikz[baseline=-.65ex]\protect\draw[thick, color=color_cvae, fill=color_cvae, mark=*, mark size=2pt, line width=1.25pt] plot[] (-.0, 0)--(-0,0);}) \textsc{cvae}                      & \cmark       & \xmark                    & \xmark                    & -                          & -                         & -                         & -                         & \citet{Sohn15}     \\
            \tikzexternalenable
            \tikzexternaldisable ({\protect\tikz[baseline=-.65ex]\protect\draw[thick, color=color_gppvae, fill=color_gppvae, mark=*, mark size=2pt, line width=1.25pt] plot[] (-.0, 0)--(-0,0);}) \textsc{gppvae}                & \cmark       & \cmark                    & \xmark                    & $\mathcal{O}(N H^2)$       & \xmark                    & \xmark                    & \xmark                    & \citet{Casale18}   \\
            \tikzexternalenable
            \tikzexternaldisable ({\protect\tikz[baseline=-.65ex]\protect\draw[thick, color=color_gpvae, fill=color_gpvae, mark=*, mark size=2pt, line width=1.25pt] plot[] (-.0, 0)--(-0,0);}) \textsc{gpvae}                   & \xmark       & \cmark                    & \xmark                    & $\mathcal{O}(N^3)$         & \cmark                    & \xmark                    & \xmark                    & \citet{Pearce19}   \\
            \tikzexternalenable
            \tikzexternaldisable ({\protect\tikz[baseline=-.65ex]\protect\draw[thick, color=color_gp_vae, fill=color_gp_vae, mark=*, mark size=2pt, line width=1.25pt] plot[] (-.0, 0)--(-0,0);}) \textsc{gp}-\textsc{vae}       & \cmark       & \cmark                    & \xmark                    & $\mathcal{O}(N)$           & \xmark                    & \xmark                    & \xmark                    & \citet{Fortuin20}  \\
            \tikzexternalenable
            \tikzexternaldisable ({\protect\tikz[baseline=-.65ex]\protect\draw[thick, color=color_sgp_vae, fill=color_sgp_vae, mark=*, mark size=2pt, line width=1.25pt] plot[] (-.0, 0)--(-0,0);}) \textsc{sgp}-\textsc{vae}    & \cmark       & \cmark                    & \xmark                    & $\mathcal{O}(B M^2 + M^3)$ & \cmark                    & \cmark                    & \xmark                    & \citet{Ashman20}   \\
            \tikzexternalenable
            \tikzexternaldisable ({\protect\tikz[baseline=-.65ex]\protect\draw[thick, color=color_svgp_vae, fill=color_svgp_vae, mark=*, mark size=2pt, line width=1.25pt] plot[] (-.0, 0)--(-0,0);}) \textsc{svgp}-\textsc{vae} & \cmark       & \cmark                    & \xmark                    & $\mathcal{O}(B M^2 + M^3)$ & \cmark                    & \cmark                    & \xmark                    & \citet{Jazbec21a}  \\
            \midrule
            \tikzexternaldisable ({\protect\tikz[baseline=-.65ex]\protect\draw[thick, color=color_bae, fill=color_bae, mark=*, mark size=2pt, line width=1.25pt] plot[] (-.0, 0)--(-0,0);}) \textsc{bae}                         & \cmark       & \xmark                    & \cmark                    & -                          & -                         & -                         & -                         & \textbf{This work} \\
            \tikzexternalenable
            \tikzexternaldisable ({\protect\tikz[baseline=-.65ex]\protect\draw[thick, color=color_gpbae, fill=color_gpbae, mark=*, mark size=2pt, line width=1.25pt] plot[] (-.0, 0)--(-0,0);}) \textsc{gp}-\textsc{bae}                   & \xmark       & \cmark                    & \cmark                    & $\mathcal{O}(N^3)$         & \cmark                    & \xmark                    & \xmark                    & \textbf{This work} \\
            \tikzexternalenable
            \tikzexternaldisable ({\protect\tikz[baseline=-.65ex]\protect\draw[thick, color=color_bsgpae, fill=color_bsgpae, mark=*, mark size=2pt, line width=1.25pt] plot[] (-.0, 0)--(-0,0);}) \textsc{sgp}-\textsc{bae}                & \cmark       & \cmark                    & \cmark                    & $\mathcal{O}(B M^2 + M^3)$ & \cmark                    & \cmark                    & \cmark                    & \textbf{This work} \\
            \bottomrule
            \tikzexternalenable
        \end{tabular}}
        
    \vspace{-3.5ex}
\end{table*}

\section{Introduction}
The problem of learning representations of data that are useful for downstream tasks is a crucial factor in the success of many machine learning applications \citep{bengio2013representation}. 
Among the numerous proposed solutions, modeling approaches that evolved from \glspl{AE} \citep{Cottrell89} are particularly appealing, as they do not require annotated data and have proven effective in unsupervised learning tasks, such as data compression and generative modeling~\citep{tomczak2022deep,yang2022introduction}.
\glspl{AE} are neural networks consisting of an encoder that maps input data to a set of lower-dimensional latent variables and a decoder that maps the latent variables back to the observations.

In applications where data is scarce or uncertainty quantification and learning interpretable representations are crucial, it is beneficial to treat these models in a Bayesian manner \citep{Mackay1992,neal1996bayesian,wilson20,Izmailov21} by imposing meaningful prior distributions over both their latent spaces as well as their parameters, i.e.,  the parameters of the encoder/decoder.
However, a fully Bayesian treatment of these models, which we refer to as \glspl{BAE}, is challenging due to the huge number of local (per-datapoint) latent variables to be inferred. This difficulty has motivated the development of \glspl{VAE} \citep{Kingma14, rezende2014stochastic} that make inference tractable by amortization. Note that \glspl{VAE} do not capture Bayesian uncertainty in the decoder parameters.

A critical limitation of standard \glspl{VAE} is that latent encodings are assumed to be factorized across latent dimensions.
However, capturing latent correlations is often necessary to model structured data.
For example, in autonomous driving or medical imaging applications, high-dimensional images are correlated in time. 
Spatio-temporal dependencies between samples are also common in environmental and life sciences datasets.

To address this limitation, several works \citep{Pearce19,Casale18} have attempted to introduce Gaussian process (\gls{GP}) priors over the latent space of \glspl{VAE} that capture correlations between pairs of latent variables through a kernel function.
While \gls{GP} priors outperform conventional priors on many tasks, they also introduce computational challenges such as $\mathcal{O}(N^{3})$ complexity for \gls{GP} inference, where $N$ is the number of data instances. 
Recently, \citet{Jazbec21a} proposed the \svgpvae model to tackle this computational issue by relying on sparse approximations which summarize the dataset into a set of points referred to as inducing points \citep{quinonero-candela05a}.

Although the \svgpvae model \citep{Jazbec21a} has achieved promising results, it has several significant drawbacks.
First, similarly to \glspl{VAE}, \svgpvae is strongly tied to a \gls{VI} formulation \citep{jordan1999introduction,zhang2018advances}, which can lead to poor approximations due to \gls{VI} making strong assumptions on both the factorization and functional form of the posterior.
Second, the \svgpvae model follows the common practice in the sparse \gls{GP} literature of optimizing the inducing points and kernel hyperparameters based on the marginal likelihood. 
However, this approach does not account for uncertainty in the inducing inputs and hyperparameters. It is well known that this can result in biased estimates and underestimated predictive uncertainties \citep{Rossi21, Lalchand22}.
In this work, we propose a novel \gls{SGPBAE} model that addresses these issues by providing scalable and flexible inference through a fully Bayesian treatment.

\vspace{-2ex}

\paragraph{Contributions.}
Specifically, \textbf{first}, we develop a fully Bayesian autoencoder (\textsc{bae}) model (\cref{sec:bae}), where we adopt a Bayesian treatment for both the local (per-datapoint) latent variables and the global decoder parameters.
This approach differs from \glspl{VAE}  in that it allows specifying any prior over the latent space while decoupling the model from the inference.
As a result, we can rely on powerful alternatives to \gls{VI} to carry out inference, and we adopt \gls{SGHMC} \citep{Cheni2014} as a scalable solution. 
To achieve this, we propose an amortized \gls{MCMC} approach for our \gls{BAE} model by using an implicit stochastic network as the encoder to learn to draw samples from the posterior of local latent variables.
Our approach addresses the prohibitively expensive inference cost induced by the local latent variables and avoids making strong assumptions on the form of the posterior.
\textbf{Second}, when imposing a \gls{GP} prior over the latent space, we propose a novel scalable \gls{SGPBAE} model (\cref{sec:bsgpae}) in which the inducing points and kernel hyperparameters of the sparse \gls{GP} prior on the latent space are treated in a fully Bayesian manner. 
This model offers attractive features such as high scalability, richer modeling capability, and improved prediction quality. 
\textbf{Third}, we extend \gls{SGPBAE} model to allow deep \gls{GP} prior  \citep{Damianou13}  and handling missing data.
To the best of our knowledge, this is the first work to consider deep \gls{GP} prior to \glspl{AE}.
\textbf{Finally}, we conduct a rigorous evaluation of our \gls{SGPBAE} model through a variety of experiments on dynamic representation learning and generative modeling.
The results demonstrate excellent performance compared to existing methods of combining \glspl{GP} and \glspl{AE} (\cref{sec:experiments}).

\section{Related Work}

\vspace{-0.5ex}

\paragraph{Autoencoders.}
\glspl{AE} \citep{Cottrell89} are a powerful framework for representation learning by projecting data into a lower-dimensional latent space using an encoder-decoder architecture.
\glspl{VAE} \citep{Kingma14, rezende2014stochastic} elegantly combines \gls{AE} with variational inference enabling the model to generate new data and allowing for the specification of any prior on the latent space. 
To improve the performance of \glspl{VAE}, recent works have attempted to employ flexible priors such as mixture distributions \citep{TomczakW18, BauerM19}, normalizing flows \citep{Chen2017}, non-parametric model \citep{NalisnickS17}, or Gaussian processes \citep{Casale18}. %
In this work, we take an orthogonal view as we consider Bayesian autoencoders, which recently have shown promising performance \citep{Tran21}, but still lack a mechanism to impose priors on the latent space.

\vspace{-2.8ex}

\paragraph{Gaussian process priors for AE models.}
The earliest attempts to combine \gls{AE} models with \glspl{GP} are the \gls{GP} prior \gls{VAE} (\gppvae) \citep{Casale18} and \gpvae \citep{Pearce19}.
Both these models are not scalable for generic kernel choices and data types.
\gppvae is restricted to a view-object \gls{GP} product kernel and employs a Taylor approximation for \glspl{GP}, while \gpvae relies on exact \gls{GP} inference.
Recently, \citet{Fortuin20} and \citet{zhu2022markovian} propose \textsc{gp}-\textsc{vae} and \textsc{m}arkovian-\textsc{gpvae}, respectively, that are indeed scalable (even in linear time complexity) by exploiting the Markov assumption but  they only work exclusively for time-series data.
Most closely to our method is the approach of \citet{Jazbec21a} (\svgpvae), \citet{Ashman20} (\sgpvae) and \citet{ramchandran21b}, where they utilize inducing point methods \citet{Titsias09a,Hensman13} to make \glspl{GP} scalable. 
However, all these methods strongly relied on \glspl{VAE} and a variational formulation for \glspl{GP}.
In this work, we take a completely different route, as we aim to treat sparse \glspl{GP} and \glspl{AE} in a fully Bayesian way while enjoying scalability thanks to recent advances in stochastic-gradient \gls{MCMC} sampling.
\cref{table:comparison} compares our proposed models with the key related works.

\section{Imposing Distributions over the Latent Space of Bayesian Autoencoders} \label{sec:bae}

\begin{figure}
    \centering
    \begin{subfigure}{0.17\textwidth}
        \centering
        \includegraphics[width=0.85\textwidth]{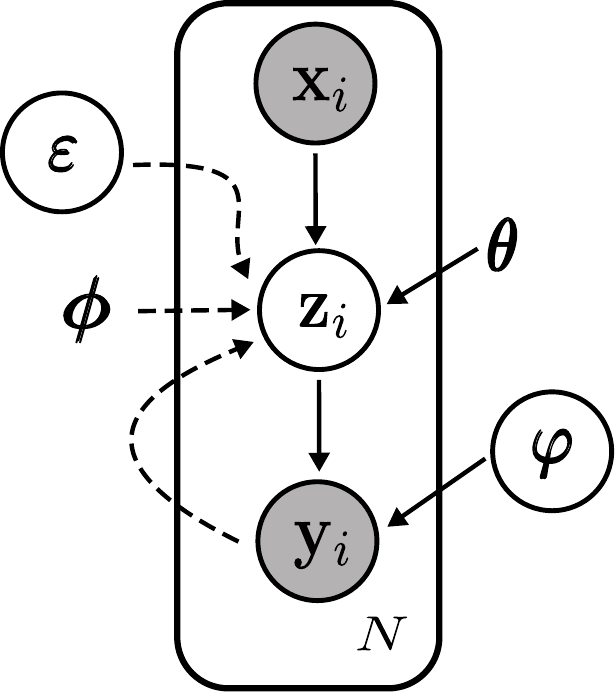}
        
        \caption{\centering \textsc{bae}. \label{fig:bae_graphical_model}}
    \end{subfigure}
    \begin{subfigure}{0.22\textwidth}
        \centering
        \includegraphics[width=0.93\textwidth]{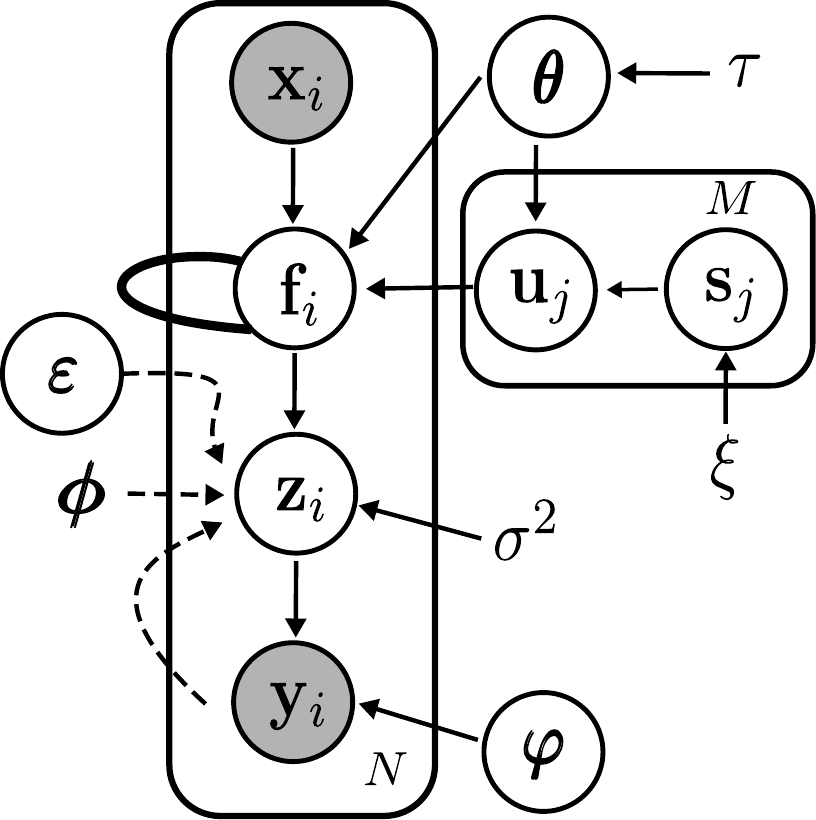}
        \caption{\centering  \textsc{sgp}-\textsc{bae}. \label{fig:bsgpae_graphical_model}}
    \end{subfigure}
    \vspace{-1ex}
    \caption{
    The graphical models of vanilla \gls{BAE} \textbf{(a)}, and the proposed \textsc{sgp}-\textsc{bae} with a fully Bayesian sparse \gls{GP} prior imposed on the latent space.
    The solid lines denote the generative part, whereas the dashed lines denote the encoding part.
    The cyclic thick line represents that the latent \gls{GP} correlates with every latent code.
    }

    \vspace{-3.9ex}
\end{figure}

\vspace{-1ex}
We are interested in unsupervised learning problems with a high-dimensional dataset consisting of $N$ data points $\mbY \stackrel{\text{def}}{=} [\mby_1, \cdots, \mby_N]^{\top} \in \mathbb{R}^{N \times P}$.
Each data point has a corresponding low-dimensional auxiliary data entry, summarized as $\mbX \stackrel{\text{def}}{=} [\mbx_1, \cdots, \mbx_N]^{\top} \in \mathbb{R}^{N \times D}$. %
For instance, $\mby_i$ could be a video frame and $\mbx_i$ the corresponding time stamp.
As another example, consider electronic health record (EHR) data, where the auxiliary data could include patient covariate information, such as age, height, weight, sex, and time since remission.
Finally, we denote by $\mbZ \stackrel{\text{def}}{=} [\mbz_1, \cdots, \mbz_N]^{\top} \in \mathbb{R}^{N \times C}$ the low-dimensional latent representation of the data, meaning that each latent variable $\mbz_i$ lives in a $C$-dimensional latent space.
We aim to train a model that is able to (1) generate $\mbY$ conditioned on $\mbX$, and (2) provide useful  and interpretable low-dimensional representations of $\mbY$.  
\vspace{-2.5ex}

\paragraph{Model setup.}
In this work, we consider a model based on \glspl{AE}, and we aim to treat this in a fully Bayesian manner.
This treatment promises improved predictions, reliable uncertainty estimates, and increased robustness under data sparsity \citep{Mackay1992, izmailov21a}.
One difficulty in doing so is that the prior distribution over the latent variables would be determined by the prior over the weights of the encoder and not a distribution of interest.
In many applications, including the ones considered here, this is undesirable, and the goal is to impose a certain prior distribution over the latent representation in a similar vein as \glspl{VAE} and their variants. 
Therefore, we propose to treat the entire \gls{AE} in a fully Bayesian manner except for the encoder, and to design the encoder in such a way that it maps data $\mby_i$ into corresponding codes $\mbz_i$ while allowing these mappings to be compatible with posterior samples over the latent codes. 
For the encoder, as we will elaborate on shortly, we will employ a so-called stochastic inference network to learn to draw posterior samples of latent variables $\mbz_i$ given high-dimensional inputs $\mby_i$, while for the decoder and the latent variables we employ scalable \gls{MCMC} techniques. 

\vspace{-2.0ex}

\paragraph{Bayesian treatment of the latent space and the decoder.} 
In order to retain a fully Bayesian treatment of the latent space and the decoder, we impose a prior $p(\mbvarphi)$ over the decoder's parameters, $\mbvarphi$.
In addition, another prior $p(\mbZ \g \mbX; \mbtheta)$ is imposed on the latent variables.
This prior is conditioned on the auxiliary data $\mbX$ and characterized by a set of hyper-parameters $\mbtheta$.
For example, one may employ an uninformative prior such as an isotropic Gaussian commonly used in standard \glspl{VAE}, but in the next section we will consider structured priors such as \glspl{GP} and their deep version as well. 
We assume that the observed data $\mbY$ is fully factorized and conditional on the latent variables $\mbZ$ and a decoder network with parameters $\mbvarphi$, i.e., $p(\mbY \g \mbZ, \mbvarphi) = \prod_{i=1}^{N} p(\mby_i \g \mbz_i, \mbvarphi)$.
The full joint distribution of the model is as follows:
\begin{align}
    \vspace{-1ex}
    p(\mbvarphi, \mbZ, \mbY \g \mbX) &=  p(\mbvarphi)  p(\mbZ \g \mbX; \mbtheta) p(\mbY \g \mbZ, \mbvarphi). \label{eq:bae_joint} %
\end{align}
We wish to infer the posterior over the latent variables and decoder parameters, which is given by Bayes' rule:
\begin{align}
    p(\mbvarphi, \mbZ \g \mbY, \mbX) = \frac{p(\mbvarphi, \mbZ, \mbY \g \mbX)}{p(\mbY \g \mbX)}, \label{eq:bae_posterior}
\end{align}
where  $p(\mbY \g \mbX) = \int p(\mbvarphi, \mbZ, \mbY \g \mbX) d\mbvarphi d\mbZ$ is the marginal likelihood.
The generative process of data samples from this \gls{BAE} model is illustrated in \cref{fig:bae_graphical_model}.

\vspace{-1ex}
Characterizing the posterior distribution over $\mbvarphi, \mbZ$ is analytically intractable and requires approximations.
Given the success of scalable \gls{MCMC} techniques to obtain samples from the posterior of model parameters in deep learning models \citep{ZhangLZCW20,Tran22}, in this work, we propose to follow this practice to obtain samples from $p(\mbvarphi, \mbZ \g \mbY, \mbX)$.
In particular, we employ \gls{SGHMC} \citep{Cheni2014}, %
which can scale up to large datasets by relying on noisy but unbiased estimates of the energy function (log-posterior) $U(\mbvarphi, \mbZ; \mbX, \mbY) \stackrel{\text{def}}{=} -\log p(\mbvarphi, \mbZ, \mbY \g \mbX)$ and without the need to evaluate the energy function over the entire data set.
More precisely, when the prior over the latent codes is fully factorized, we can approximate this energy function using mini-batches of size $B$ as follows:
\begin{align}
    {U}(\mbvarphi,  \mbZ; \mbX, \mbY) & \approx \tilde{U}(\mbvarphi,  \mbZ; \mbX, \mbY) = -\log p(\mbvarphi) - \label{eq:energy_bae} \\ 
    - \frac{N}{B} \sum_{i \in \cI_B} &\big[ \log p(\mbz_i \g \mbx_i; \mbtheta) + \log p(\mby_i \g \mbz_i, \mbvarphi) \big] \nonumber
\end{align}
where $\cI_B$ is a set of $B$ random indices.
The exact procedure for generating samples from the posterior over $\mbvarphi, \mbZ$ using \gls{SGHMC} can be found in \cref{sec:sghmc}.

\vspace{-2.0ex}

\paragraph{Encoder as a stochastic inference network.} \gls{SGHMC} can be challenging to implement on probabilistic models with many latent variables due to the high computational burden of iteratively refining the approximate posterior for each latent variable. 
Additionally, it can be difficult to evolve the latent variables for each new test sample.
To address these challenges, we propose using a stochastic neural network as an inference network to efficiently generate latent codes similar to those generated by the posterior distribution, inspired by amortized inference techniques \citep{Kingma14, wang2016learning, FengWL17, Shi19a} and \gls{MCMC} distillation \citep{Balan2015, Wang18i, li2017approximate}.

\vspace{-0.3ex}

More specifically, instead of storing every latent code, we use an inference network $\mbz_i = f_{\mbphi}(\mby_i; \mathbold{\varepsilon})$ with parameters $\mbphi$ that generates a corresponding latent code $\mbz_i$ given an input $\mby_i$ and a random seed $\mathbold{\varepsilon}$.
The random seed $\mathbold{\varepsilon}$ is drawn from a distribution $q(\mathbold{\varepsilon})$ that is easy to sample from, such as a uniform or standard Gaussian distribution.
The inference network $f_{\phi}$ serves as an encoder by generating posterior samples of the latent code $\mbz_i$ given the observed input $\mby_i$. 
Our approach using this encoder differs from that of \glspl{VAE}, as we do not make any assumptions about the form of the posterior of the latent variables $\mbz_i$.

\vspace{-0.3ex}

We incrementally refine the encoder $f_{\mbphi}$ such that its outputs mimic the \gls{SGHMC} dynamics.
Specifically, after every $K$ iterations of sampling the decoder parameters and the latent codes using \gls{SGHMC}, we adjust the encoder parameters $\mbphi$ based on the following objective:
\begin{align}
    \cL\left(\{\mbz_i, \mby_i\}_{i\in \cI_B} ; \mbphi\right) \stackrel{\text{def}}{=} \sum_{i \in \cI_{B}} \Big\| f_{\mbphi}(\mby_i; \boldsymbol{\varepsilon}_i ) - \mbz_i^{(k)} \Big\|_2^2, \label{eq:encoder_objective}
\end{align}
 where $\mbz_i^{(k)}$ is the $k$-th posterior sample from \gls{SGHMC} of the latent variable $\mbz_i$, and it is used as label to update $\mbphi$.
As the analytic solution of \cref{eq:encoder_objective} is intractable, we perform $J$ steps of gradient descent to update $\mbphi$ using an optimizer such as Adam \citep{jlb2015adam}.
The inference procedure for \glspl{BAE} is described in \cref{alg:bae_sampling}.

\vspace{-2ex}

{
\begin{algorithm}[]
    
    \caption{Inference for \glspl{BAE} with \gls{SGHMC}}
    \label{alg:bae_sampling}

    \KwInput{Dataset $\{\mbX, \mbY\}$, mini-batch size $B$, \# \gls{SGHMC} iterations $K$, \# encoder interations $J$}
    Initialize the autoencoder parameters $\mbphi$ and $\mbvarphi$ \\
    \While{$\mbvarphi$, $\mbZ$ have not converged}{
        Sample a mini-batch of $B$ random indices $\cI_B$ \\
        Sample random seed $\{ \boldsymbol{\varepsilon}_i \}_{i=1}^{B} \sim q(\mathbold{\varepsilon})$  \\
        Initialize the latent codes from the encoder $\{\mbz_i\}_{i=1}^{B} = f_{\mbphi}(\{\mby_i\}_{i \in \cI_B}, \{ \boldsymbol{\varepsilon}_i \}_{i=1}^{B})$ \\
        \For{K iterations}{
            Compute energy function $\tilde{U}$ using \cref{eq:energy_bae} \label{line:energy_computing} \\
            Sample from posterior $p(\mbvarphi, \mbZ \g \mbY, \mbX)$: \label{line:posterior_sampling}
            $\mbvarphi, \{\mbz\}_{i=1}^{B} \leftarrow$ \texttt{SGHMC}($\mbvarphi, \{\mbz\}_{i=1}^{B}; \nabla_{\mbvarphi, \mbz}\tilde{U}$)
        }    
        \For{J iterations}{
            Compute objective function $\cL$ using \cref{eq:encoder_objective} \\
            Update encoder: $\mbphi \leftarrow \texttt{Optimizer}(\mbphi; \nabla_{\mbphi}\cL)$
        }
    }

\end{algorithm}

}

\section{Scalable Gaussian Process Prior for Bayesian Autoencoders} \label{sec:bsgpae}

In the previous section, we introduced our novel version of a \gls{BAE} where we imposed a simple fully-factorized prior over the latent space, such as isotropic Gaussians.
However, in many applications, such priors are incapable of appropriately modeling the correlation nature of the data.
For example, it is sensible to model structured data evolving over time with a \gls{BAE} with a prior over the latent space in the form of a \gls{GP} with the auxiliary data as input.
In this section, we consider these scenarios precisely by introducing \gls{GP} priors in the latent space, which allows us to model sample covariances as a function of the auxiliary data.
We then discuss the scalability issues induced by the use of \gls{GP} priors, and we propose \glsreset{BSGPAE} \gls{SGPBAE} where we recover scalability thanks to sparse approximations to \glspl{GP} \citep{quinonero-candela05a, Rossi21}.
In this model, we carry out fully Bayesian inference of the decoder, as well as the inducing inputs and covariance hyperparameters of the sparse \glspl{GP}, while we optimize the stochastic inference network implementing the encoder.

\vspace{-1.5ex}
\subsection{Gaussian process prior}

\vspace{-0.5ex}

We assume $C$ independent latent functions $f^{[1]}, ..., f^{[C]}$, which results in each $\mbz_i$ being evaluated at the corresponding $\mbx_i$, i.e., $\mbz_i = \left[f^{[1]}(\mbx_i), \cdots, f^{[C]}(\mbx_i)\right]$.
We assume that each function is drawn from a zero-mean \gls{GP} prior with a covariance function $\kappa(\mbx, \mbx'; \mbtheta)$: 
\begin{align}
    p(\mbZ \g \mbX; \mbtheta) = \prod_{c=1}^{C} \cN\left(\mbz^{[c]}_{1:N} \g \mathbold{0}, \mbK_{\mbx \mbx \g \mbtheta}\right),
\end{align}
where the $c$-th latent channel of all latent variables, $\mbz^{[c]}_{1:N} \in \mathbb{R}^{N}$ (the $c$-th column of $\mbZ$), has a correlated Gaussian prior with covariance $\mbK_{\mbx \mbx \g \mbtheta} \in \mathbb{R}^{N \times N}$ obtained by evaluating $\kappa(\mbx_i, \mbx_j; \mbtheta)$ over all input pairs of $\mbX$.
Here, the latent function values are informed by all $\mby$ values according to the covariance of the corresponding auxiliary input $\mbx$.
One can recover to the fully factorized $\cN(\mathbold{0}, \mbI)$ prior on the latent space by simply setting $\mbK_{\mbx\mbx \g \mbtheta} = \mbI$.

This \gls{GP} prior over the latent space of \glspl{BAE} introduces fundamental scalability issues.
First, we have to compute the inverse and log-determinant of the kernel matrix $\mbK_{\mbx \mbx \g \mbtheta}$, which results in $\mathcal{O}(N^{3})$ time complexity.
This is only possible when $N$ is  of moderate size.
Second, it is impossible to employ a mini-batching inference method like \gls{SGHMC} since the energy function $U(\mbvarphi, \mbZ; \mbX, \mbY)$ does not decompose as a sum over all the observations.

\vspace{-1.5ex}

\subsection{Bayesian sparse Gaussian processes}

\vspace{-0.5ex}

In order to keep the notation uncluttered, we focus on a single channel and suppress the superscript index $c$.
Given a set of latent function evaluations over the dataset, $\mbf = [f_1, \cdots, f_N]^{\top}$, we assume that the latent codes are stochastic realizations based on $\mbf$ and additive Gaussian noise i.e., $\cN(\mbZ \g \mbf, \sigma^2 \mbI)$.
Sparse \glspl{GP} \citep{quinonero-candela05a} are a family of approximate models that address the scalability problem by introducing a set of $M \ll N$ inducing points $\mbu = (u_1, \cdots, u_M)$ at corresponding inducing inputs $\mbS = \{ \mbs_1, \cdots, \mbs_M \}$ such that $u_i = f(\mbs_i)$.
We assume that these inducing variables follow the same \gls{GP} as the original process, resulting in the following joint prior: %
\begin{align}
    p(\mbf, \mbu) = \cN \bigg( \mathbold{0},
        \begin{bmatrix}
        \mbK_{\mbX \mbX \g \mbtheta} & \mbK_{\mbX \mbS \g \mbtheta} \\
        \mbK_{\mbS \mbX \g \mbtheta} & \mbK_{\mbS \mbS \g \mbtheta}
        \end{bmatrix} \bigg),
\end{align}
where the covariance matrices $\mbK_{\mbS \mbS \g \mbtheta}$ and $\mbK_{\mbX \mbS \g \mbtheta}$ are computed between the elements in $\mbS$ and $\{ \mbX, \mbS \}$, respectively.

\vspace{-1ex}

\paragraph{Fully Bayesian sparse \glspl{GP}.}
The fully Bayesian treatment of sparse \glspl{GP} requires priors $p_{\tau}(\mbtheta)$ and $p_{\xi}(\mbS)$ over covariance hyper-parameters and inducing inputs, respectively, with $\tau$ and $\xi$ as prior hyper-parameters.
With these assumptions, we term this model as Bayesian sparse Gaussian process autoencoder (\gls{SGPBAE}), and the corresponding generative model is illustrated in \cref{fig:bsgpae_graphical_model}. 

By defining $\mbPsi \stackrel{\text{def}}{=} \{\mbvarphi, \mbu, \mbS, \mbtheta \}$, we can rewrite the full joint distribution of parameters in \gls{SGPBAE}:
\begin{align}
    & p(\mbPsi,\mbf, \mbZ, \mbY \g \mbX) = \\
    &= \hspace{-4ex} \underbrace{p(\mbPsi)}_{\substack{\text{Priors on inducing} \\ \text{inputs \& variables, decoder}}} \hspace{-2ex} \underbrace{p(\mbf \g \mbu, \mbX, \mbS, \mbtheta) p(\mbZ \g \mbf; \sigma^2 \mbI)}_{\text{Sparse GP prior on latent space}} \underbrace{p(\mbY \g \mbZ, \mbvarphi)}_{\substack{\text{Likelihood of} \\ \text{observed data}}} \nonumber,
    \end{align}
where $p(\mbPsi) = p(\mbvarphi) p_{\tau}(\mbtheta) p_{\xi}(\mbS) p(\mbu \g \mbS, \mbtheta)$.
Here, $p(\mbu \g \mbS, \mbtheta) = \cN(\boldsymbol{0, \mbK_{\mbS \mbS \g \mbtheta} })$, and $p(\mbf \g \mbu, \mbX, \mbS, \mbtheta) = \cN(\mbK_{\mbX \mbS \g \mbtheta} \mbK_{\mbS \mbS \g \mbtheta}^{-1} \mbu, \mbK_{\mbX \mbX \g \mbtheta} - \mbK_{\mbX \mbS \g \mbtheta} \mbK_{\mbS \mbS \g \mbtheta}^{-1} \mbK_{\mbS \mbX \g \mbtheta} )$.
We assume a factorization $p(\mbZ \g \mbf; \sigma^2 \mbI) = \prod_{i=1}^{N}p(\mbz_i \g f_i; \sigma^2)$ and make no further assumtions about the other distribution.

\vspace{-1ex}

\paragraph{Scalable inference objective.} We wish to infer the set of variables $\mbPsi$ and the latent codes $\mbZ$.
To do so, we have to marginalize out the latent process $\mbf$ from the full joint distribution above.
In particular, we have:
\begin{align}
    & \log   p(   \mbPsi, \mbZ, \mbY \g \mbX) =  \log p(\mbPsi) + \\  \label{eq:bsgpae_sampling_obj}
   & + \log \int p(\mbf \g \mbPsi, \mbX) p(\mbZ \g \mbf, \sigma^2 \mbI) d\mbf + \log p(\mbY \g \mbZ, \mbvarphi). \nonumber
\end{align}
This objective should be decomposed over observations to sample from the posterior over all the latent variables using a scalable method such as \gls{SGHMC}.
As discussed by \citet{Rossi21}, this can be done effectively by imposing independence in the conditional distribution \citep{Snelson06}, i.e., by parameterizing $p(\mbf \g \mbPsi, \mbX) = \cN \big( \mbK_{\mbx \mbs \g \mbtheta} \mbK_{\mbs \mbs \g \mbtheta}^{-1} \mbu, \diag \big[ \mbK_{\mbx \mbx \g \mbtheta} - \mbK_{\mbx \mbs \g \mbtheta} \mbK_{\mbs \mbs \g \mbtheta}^{-1} \mbK_{\mbs \mbx  \g \mbtheta} \big] \big)$.
With this approximation, the log joint marginal becomes as follows:
\begin{align}
    \log &  p( \mbPsi, \mbZ, \mbY \g \mbX) \approx \log p(\mbPsi) + \\ \nonumber \label{eq:energy_bsgpae}
    \vspace{-3ex} + \sum_{n=1}^{N} & \bigg\{ \log \mathbb{E}_{p(f_n \g \mbPsi, \mbX)}[p(\mbz_n \g f_n; \sigma^2)] +  \log p(\mby_n \g \mbz_n, \mbvarphi) \bigg\} \\
    & \stackrel{\text{def}}{=} -U(\mbPsi, \mbZ;\mbX, \mbY). \nonumber
 \end{align}
\vspace{-0.1ex}
We can now carry out inference for this \gls{SGPBAE} model by plugging a mini-batching approximation of this energy function into \cref{line:energy_computing} and \cref{line:posterior_sampling} of \cref{alg:bae_sampling}.
By using this sparse approximation, we reduce the computational complexity of evaluating the \gls{GP} prior down to $\mathcal{O}(B M^2)$, and \gls{SGPBAE} can be readily applied to a generic dataset and arbitrary \gls{GP} kernel function.

\vspace{-2ex}

\paragraph{Extension to deep Gaussian processes.} \label{sec:deep_gp} 
We can easily extend \gls{SGPBAE} to deep Gaussian process priors \citep{Damianou13} to model much more complex functions in the latent space of \glspl{BAE}.
To the best of our knowledge, the use of deep \glspl{GP} has not been considered in previous work. 
We assume a deep Gaussian process prior $f^{(L)} \circ f^{(L-1)} \circ \cdots \circ f^{(1)}$, where each $f^{(l)}$ is a \gls{GP}. 
Each layer is associated with a set of kernel hyperparameters $\mbtheta^{(l)}$, inducing inputs $\mbS^{(l)}$ and inducing variables $\mbu^{(l)}$.%
The set of variables to be inferred is $\mbPsi = \{\mbvarphi\} \cup \{ \mbu^{(l)}, \mbS^{(l)}, \mbtheta^{(l)}\}_{l=1}^{L}$.
The joint distribution is as follows:
\begin{align}
     p(\mbPsi, & \{\mbf^{(l)}\}_{l=1}^{L}, \mbZ, \mbY \g \mbX) = \\
     = p(\mbPsi)& \underbrace{\prod_{l=1}^{L}  p(\mbf^{(l)} \g \mbf^{(l-1)}, \mbPsi)  p(\mbZ \g \mbf^{(L)}; \sigma^2 \mbI)}_{\text{Deep GP prior }} p(\mbY \g \mbZ, \mbvarphi), \nonumber
\end{align}
\vspace{-0.1ex}
where we omit the dependency on $\mbX$ for notational brevity.
To perform inference, the hidden layers $\mbf^{(l)}$ have to be marginalized and propagated up to the final layer $L$ \citep{Salimbeni17}.
The marginalization can be approximated by quadrature \citep{HensmanMG15} or through Monte Carlo sampling \citep{Edwin17}.
Detailed derivations of this extension can be found in \cref{sec:deep_gp_appendix}.

\begin{figure*}[h]
    \begin{minipage}{0.7\textwidth}
        {
    \tikzexternaldisable
    \centering
    \normalsize
    \setlength{\figurewidth}{3.20cm}
    \setlength{\figureheight}{3.20cm}
    \setlength{\tabcolsep}{0.5pt}

    \captionof{table}{
        Reconstructions of the latent trajectories of moving ball.
        In the first column, frames of each test video are overlayed and shaded by time.
        Ground truth trajectories are illustrated in \textcolor{color_orange_trajectory}{\textbf{orange}}, while predicted trajectories are depicted in \textcolor{color_blue_trajectory}{\textbf{blue}}.
        We use $M=10$ inducing points for the methods employed with sparse \glspl{GP}.
        \label{tab:moving_ball}
    }
    \vspace{-1.5ex}
    \small
    \begin{center}
        \scalebox{.85}
        {
        \begin{tabular}{ccccccc}
            \toprule
            \textsc{gt video}                                                                                                                                                                                                              &
            \tikzexternaldisable ({\protect\tikz[baseline=-.65ex]\protect\draw[thick, color=color_vae, fill=color_vae, mark=*, mark size=2pt, line width=1.25pt] plot[] (-.0, 0)--(-0,0);}) \tikzexternalenable \textsc{vae}               &
            \tikzexternaldisable ({\protect\tikz[baseline=-.65ex]\protect\draw[thick, color=color_gpvae, fill=color_gpvae, mark=*, mark size=2pt, line width=1.25pt] plot[] (-.0, 0)--(-0,0);}) \tikzexternalenable \textsc{gpvae}             &
            \tikzexternaldisable ({\protect\tikz[baseline=-.65ex]\protect\draw[thick, color=color_svgp_vae, fill=color_svgp_vae, mark=*, mark size=2pt, line width=1.25pt] plot[] (-.0, 0)--(-0,0);}) \tikzexternalenable \textsc{svgp}-\textsc{vae} &
            \tikzexternaldisable ({\protect\tikz[baseline=-.65ex]\protect\draw[thick, color=color_bae, fill=color_bae, mark=*, mark size=2pt, line width=1.25pt] plot[] (-.0, 0)--(-0,0);}) \tikzexternalenable \textsc{bae}               &
            \tikzexternaldisable ({\protect\tikz[baseline=-.65ex]\protect\draw[thick, color=color_gpbae, fill=color_gpbae, mark=*, mark size=2pt, line width=1.25pt] plot[] (-.0, 0)--(-0,0);}) \tikzexternalenable \textsc{gp-bae}             &
            \tikzexternaldisable ({\protect\tikz[baseline=-.65ex]\protect\draw[thick, color=color_bsgpae, fill=color_bsgpae, mark=*, mark size=2pt, line width=1.25pt] plot[] (-.0, 0)--(-0,0);}) \tikzexternalenable \textsc{sgp}-\textsc{bae}                                                                                                                                                                                                                                                                  \\
            \midrule                                                                                                                                                                                                                                                                                                                                                                                                                                                                             \\ [-18.5pt]
            \raisebox{-7.7pt}{\includegraphics[clip,width=0.179\linewidth]{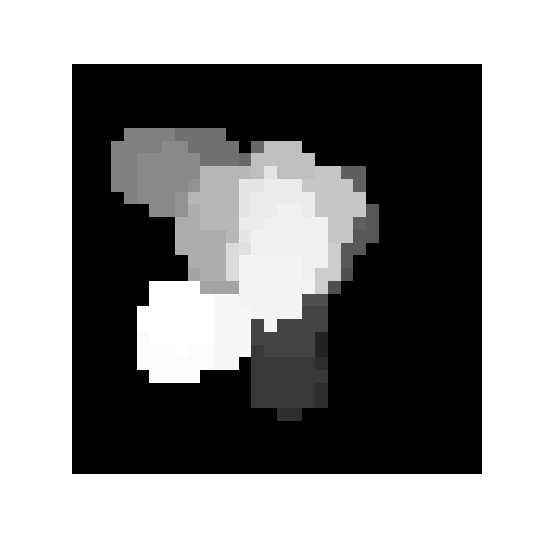}}                                                                                                                                        & \begin{tikzpicture}

\definecolor{dodgerblue}{RGB}{30,144,255}
\definecolor{orangered}{RGB}{255,69,0}

\begin{axis}[
height=\figureheight,
major tick length=1ex,
tick pos=left,
ticks=none,
width=\figurewidth,
xmin=-2.3, xmax=2.3,
xticklabels={},
ymin=-2.3, ymax=2.3,
yticklabels={}
]
\addplot [thick, dodgerblue, opacity=0.7]
table {%
0.0517649599523009 0.0285932281730593
0.0990116974783536 0.0569416236817002
0.0346024413861383 0.0453272951020148
-0.0522942401979837 0.00554039716191633
0.11800478850776 -0.00318248191200132
0.965851042902721 -0.0330930326619268
0.818202241806924 0.00155376240948252
0.387889355953792 -0.00682663814422729
-0.626731024687862 0.210658453654565
-1.1849398939669 0.122945360127521
-1.28595668270709 0.125064988096423
-0.917553029316769 0.111959597043721
-0.689076646820805 -0.0148715320282375
-0.825314604879409 -0.0560343654958703
-0.950378245667259 -0.0563653322544829
-0.908496600432606 -0.0551884086086089
-0.354844602330183 0.015821566458096
0.448117154829254 -0.0121061019647637
0.757429361464174 -0.016669411854727
0.00796269796452392 -0.0285689782473691
-0.315917249228003 -0.0579010074546814
-0.503816231711394 -0.112162505648098
-0.0307160117475193 -0.0897877413811025
0.217686115036832 -0.058336380121751
0.216541032257265 -0.0717585834378798
0.00365158361683432 -0.0488604145389114
-0.314335223660021 -0.0565846369324777
-0.838463776760004 -0.0698120984298732
-1.06478994560627 0.00569989928690535
-1.77381732539401 -0.0466167321526442
};
\addplot [thick, orangered, opacity=0.7]
table {%
0.130618932297751 -0.883857092654764
0.123250835707235 -1.43454922013419
0.0162579352490851 -1.39750926524949
0.0622384237180545 -0.702845752551685
0.370445666897549 -0.0144643896060901
0.684044312145529 0.412192526549869
0.617954618380314 0.759570615444638
-0.0504062126085779 1.04847213215793
-0.982232076662367 1.19582837002712
-1.57319392041863 1.21838160119205
-1.63391533523543 1.16243230281669
-1.33702206505331 0.979000176700668
-0.893121922489428 0.563976776404871
-0.594079752937189 0.104293180933575
-0.653853094763873 -0.00779353931726327
-0.841062765579468 0.289513692216213
-0.627621036134081 0.756994929945726
0.049903944021443 0.99002199289291
0.493841004149613 0.644751567329222
0.26286022897301 -0.00143741670175838
-0.152587727706359 -0.30022800485021
-0.227980652364855 -0.049422425088862
-0.149156443903312 0.394307020807417
-0.133900236784391 0.628220199952433
-0.00820817944910509 0.534865606174796
0.12117593877965 0.191756281066145
-0.156071702460425 -0.318895437678216
-0.825222514060837 -0.850843900932538
-1.31721820450768 -1.04286875231486
-1.2759501136611 -0.639591020706136
};
\addplot [thick, orangered, opacity=0.7, mark=*, mark size=1.5, mark options={solid}]
table {%
0.130618932297751 -0.883857092654764
};
\addplot [thick, orangered, opacity=0.7, mark=triangle*, mark size=1.5, mark options={solid}]
table {%
-1.2759501136611 -0.639591020706136
};
\addplot [thick, dodgerblue, opacity=0.7, mark=*, mark size=1.5, mark options={solid}]
table {%
0.0517649599523009 0.0285932281730593
};
\addplot [thick, dodgerblue, opacity=0.7, mark=triangle*, mark size=1.5, mark options={solid}]
table {%
-1.77381732539401 -0.0466167321526442
};
\end{axis}

\end{tikzpicture} & \begin{tikzpicture}

\definecolor{dodgerblue}{RGB}{30,144,255}
\definecolor{orangered}{RGB}{255,69,0}

\begin{axis}[
height=\figureheight,
major tick length=1ex,
tick pos=left,
ticks=none,
width=\figurewidth,
xmin=-2.3, xmax=2.3,
xticklabels={},
ymin=-2.3, ymax=2.3,
yticklabels={}
]
\addplot [thick, dodgerblue, opacity=0.7]
table {%
0.293737853146464 -0.566854975750814
0.425222842154464 -1.10145249449696
0.334206791409773 -0.956368043876566
0.229063469996044 -0.452592138697001
0.330941200950125 -0.0671890532763684
0.50765156586547 0.11998063798695
0.339979870637268 0.263868166383813
-0.346346759848253 0.477474188034746
-1.21444267821981 0.74828832377565
-1.79534207812908 0.929082058258608
-1.86775732460027 0.93909912427949
-1.48756877718378 0.805198408954554
-0.89255268317039 0.494666689981513
-0.446649749235556 0.13028502826467
-0.461659095926395 0.0733068580745824
-0.765426505311216 0.33063785953735
-0.750694141004238 0.49287701038128
-0.215344378942345 0.425238766421084
0.274981777619383 0.229897311094245
0.239553127793491 -0.0339959322803801
-0.0329730420758909 -0.145644128218279
-0.144739674167489 0.00948063984416497
-0.20619065626513 0.221103592295935
-0.289873634796777 0.319346002856126
-0.163361178628802 0.264851690098138
0.077225752739437 0.0784424625835577
-0.0279390403890921 -0.153545869818941
-0.559156573472762 -0.371885520109231
-1.03840240687588 -0.432888215695626
-0.975669833955463 -0.073381097204086
};
\addplot [thick, orangered, opacity=0.7]
table {%
0.130618932297751 -0.883857092654764
0.123250835707235 -1.43454922013419
0.0162579352490851 -1.39750926524949
0.0622384237180545 -0.702845752551685
0.370445666897549 -0.0144643896060901
0.684044312145529 0.412192526549869
0.617954618380314 0.759570615444638
-0.0504062126085779 1.04847213215793
-0.982232076662367 1.19582837002712
-1.57319392041863 1.21838160119205
-1.63391533523543 1.16243230281669
-1.33702206505331 0.979000176700668
-0.893121922489428 0.563976776404871
-0.594079752937189 0.104293180933575
-0.653853094763873 -0.00779353931726327
-0.841062765579468 0.289513692216213
-0.627621036134081 0.756994929945726
0.049903944021443 0.99002199289291
0.493841004149613 0.644751567329222
0.26286022897301 -0.00143741670175838
-0.152587727706359 -0.30022800485021
-0.227980652364855 -0.049422425088862
-0.149156443903312 0.394307020807417
-0.133900236784391 0.628220199952433
-0.00820817944910509 0.534865606174796
0.12117593877965 0.191756281066145
-0.156071702460425 -0.318895437678216
-0.825222514060837 -0.850843900932538
-1.31721820450768 -1.04286875231486
-1.2759501136611 -0.639591020706136
};
\addplot [thick, orangered, opacity=0.7, mark=*, mark size=1.5, mark options={solid}]
table {%
0.130618932297751 -0.883857092654764
};
\addplot [thick, orangered, opacity=0.7, mark=triangle*, mark size=1.5, mark options={solid}]
table {%
-1.2759501136611 -0.639591020706136
};
\addplot [thick, dodgerblue, opacity=0.7, mark=*, mark size=1.5, mark options={solid}]
table {%
0.293737853146464 -0.566854975750814
};
\addplot [thick, dodgerblue, opacity=0.7, mark=triangle*, mark size=1.5, mark options={solid}]
table {%
-0.975669833955463 -0.073381097204086
};
\end{axis}

\end{tikzpicture} & \begin{tikzpicture}

\definecolor{dodgerblue}{RGB}{30,144,255}
\definecolor{orangered}{RGB}{255,69,0}

\begin{axis}[
height=\figureheight,
major tick length=1ex,
tick pos=left,
ticks=none,
width=\figurewidth,
xmin=-2.3, xmax=2.3,
xticklabels={},
ymin=-2.3, ymax=2.3,
yticklabels={}
]
\addplot [thick, dodgerblue, opacity=0.7]
table {%
-0.290817920105179 -0.398414612873631
-0.241944872029967 -0.298130646353328
-0.124621181851899 -0.111792447779476
0.056228745210899 0.151822287227376
0.293883495970414 0.480196911770026
0.544691941890223 0.813317512356577
0.708386459203216 1.01380468122544
0.691635821638468 0.953630458599941
0.496889948532307 0.648139945045266
0.214822842934322 0.252581991708029
-0.0653716021585949 -0.0791087954727996
-0.297570113786131 -0.285526561829519
-0.440309983009311 -0.361762950700675
-0.439065484318406 -0.303581358742884
-0.280808941580971 -0.120274348451069
-0.0395614911812895 0.12833927212359
0.168676741790595 0.352717711578313
0.276162306513506 0.499191999485689
0.294348793840704 0.570601283772663
0.268161473533692 0.583723260909126
0.226451106870632 0.543259095495299
0.171464058760151 0.45718367985129
0.0859062002629928 0.3368049658055
-0.0532877684477675 0.181882314509602
-0.245831756907684 -0.00526219590582094
-0.451021968999713 -0.20171599153482
-0.616474474834135 -0.380835282728057
-0.718025302505098 -0.529715723528285
-0.752404675115594 -0.633325782082873
-0.708350728672966 -0.663494233755839
};
\addplot [thick, orangered, opacity=0.7]
table {%
0.130618932297751 -0.883857092654764
0.123250835707235 -1.43454922013419
0.0162579352490851 -1.39750926524949
0.0622384237180545 -0.702845752551685
0.370445666897549 -0.0144643896060901
0.684044312145529 0.412192526549869
0.617954618380314 0.759570615444638
-0.0504062126085779 1.04847213215793
-0.982232076662367 1.19582837002712
-1.57319392041863 1.21838160119205
-1.63391533523543 1.16243230281669
-1.33702206505331 0.979000176700668
-0.893121922489428 0.563976776404871
-0.594079752937189 0.104293180933575
-0.653853094763873 -0.00779353931726327
-0.841062765579468 0.289513692216213
-0.627621036134081 0.756994929945726
0.049903944021443 0.99002199289291
0.493841004149613 0.644751567329222
0.26286022897301 -0.00143741670175838
-0.152587727706359 -0.30022800485021
-0.227980652364855 -0.049422425088862
-0.149156443903312 0.394307020807417
-0.133900236784391 0.628220199952433
-0.00820817944910509 0.534865606174796
0.12117593877965 0.191756281066145
-0.156071702460425 -0.318895437678216
-0.825222514060837 -0.850843900932538
-1.31721820450768 -1.04286875231486
-1.2759501136611 -0.639591020706136
};
\addplot [thick, orangered, opacity=0.7, mark=*, mark size=1.5, mark options={solid}]
table {%
0.130618932297751 -0.883857092654764
};
\addplot [thick, orangered, opacity=0.7, mark=triangle*, mark size=1.5, mark options={solid}]
table {%
-1.2759501136611 -0.639591020706136
};
\addplot [thick, dodgerblue, opacity=0.7, mark=*, mark size=1.5, mark options={solid}]
table {%
-0.290817920105179 -0.398414612873631
};
\addplot [thick, dodgerblue, opacity=0.7, mark=triangle*, mark size=1.5, mark options={solid}]
table {%
-0.708350728672966 -0.663494233755839
};
\end{axis}

\end{tikzpicture} & \begin{tikzpicture}

\definecolor{dodgerblue}{RGB}{30,144,255}
\definecolor{orangered}{RGB}{255,69,0}

\begin{axis}[
height=\figureheight,
major tick length=1ex,
tick pos=left,
ticks=none,
width=\figurewidth,
xmin=-2.3, xmax=2.3,
xticklabels={},
ymin=-2.3, ymax=2.3,
yticklabels={}
]
\addplot [thick, dodgerblue, opacity=0.7]
table {%
0.181723960504513 -0.641627092183683
0.521468525562838 -0.937953956259719
0.48874500233883 -0.842028581988795
0.0334445301328448 -0.516074812579798
-0.203568080854372 -0.545133114147856
0.143662895775429 0.164875177707204
0.148472210714817 0.262758657431844
-0.188145469933885 0.648902697934559
-1.12810723226691 1.33532285504756
-1.02555730823058 1.5687212991795
-1.01435056367708 1.56347217836609
-0.863763095761288 1.16088567060198
-0.627068772703067 0.640642220796574
-0.374477049419965 0.238276913832198
-0.282635136857592 0.139348472292661
-0.47271600163156 0.472695844266911
-0.695250261250971 0.557761019389681
-0.130350829151644 0.551648210793035
0.103662191586893 0.248513914377288
-0.307225026001573 -0.369394225247731
-0.196391732505264 -0.182686607849144
-0.306120602867041 -0.0475335444525172
-0.518057582501669 0.158340296705144
-0.642070890275148 0.24944105167134
-0.655081824041148 0.226680673721801
-0.511508660136505 -0.0689487738141483
-0.105620798204503 -0.30038167900394
0.034941766200753 -0.0890573604228316
0.163963410202991 -0.146823176976075
-0.0363305620667563 0.129669248882643
};
\addplot [thick, orangered, opacity=0.7]
table {%
0.130618929862976 -0.883857071399689
0.123250834643841 -1.43454921245575
0.0162579361349344 -1.39750921726227
0.0622384250164032 -0.702845752239227
0.370445668697357 -0.0144643895328045
0.684044301509857 0.412192523479462
0.617954611778259 0.759570598602295
-0.0504062138497829 1.0484721660614
-0.982232093811035 1.19582831859589
-1.57319390773773 1.21838164329529
-1.63391530513763 1.16243231296539
-1.33702206611633 0.979000151157379
-0.893121898174286 0.563976764678955
-0.594079732894897 0.104293182492256
-0.653853118419647 -0.0077935392037034
-0.841062784194946 0.289513677358627
-0.627621054649353 0.756994903087616
0.0499039441347122 0.990022003650665
0.493840992450714 0.64475154876709
0.262860238552094 -0.00143741664942354
-0.152587726712227 -0.300227999687195
-0.22798065841198 -0.0494224242866039
-0.149156451225281 0.394307017326355
-0.133900240063667 0.628220200538635
-0.00820817984640598 0.534865617752075
0.121175937354565 0.191756278276443
-0.156071707606316 -0.318895429372787
-0.825222492218018 -0.850843906402588
-1.31721818447113 -1.04286873340607
-1.27595007419586 -0.639591038227081
};
\addplot [thick, orangered, opacity=0.7, mark=*, mark size=1.5, mark options={solid}]
table {%
0.130618929862976 -0.883857071399689
};
\addplot [thick, orangered, opacity=0.7, mark=triangle*, mark size=1.5, mark options={solid}]
table {%
-1.27595007419586 -0.639591038227081
};
\addplot [thick, dodgerblue, opacity=0.7, mark=*, mark size=1.5, mark options={solid}]
table {%
0.181723960504513 -0.641627092183683
};
\addplot [thick, dodgerblue, opacity=0.7, mark=triangle*, mark size=1.5, mark options={solid}]
table {%
-0.0363305620667563 0.129669248882643
};
\end{axis}

\end{tikzpicture} & \begin{tikzpicture}

\definecolor{dodgerblue}{RGB}{30,144,255}
\definecolor{orangered}{RGB}{255,69,0}

\begin{axis}[
height=\figureheight,
major tick length=1ex,
tick pos=left,
ticks=none,
width=\figurewidth,
xmin=-2.3, xmax=2.3,
xticklabels={},
ymin=-2.3, ymax=2.3,
yticklabels={}
]
\addplot [thick, dodgerblue, opacity=0.7]
table {%
0.0606423866340622 -0.681447595736459
0.0129968684896044 -1.27188500493285
-0.032744415470314 -1.17242471124706
0.02438280982792 -0.481858297203271
0.26626027672959 0.0360124159243716
0.525263864313143 0.360430397963051
0.512205201160585 0.584614580959364
-0.106691516592438 0.863674363429419
-0.905611862863264 1.03747755516735
-1.78311760930563 1.34172425283436
-1.92903310144283 1.45327573822213
-1.2815587000767 0.878094075547768
-0.821258060134943 0.496189543485602
-0.486821058406336 0.126901311836277
-0.583189903877835 0.0442534231094629
-0.749597965164525 0.275793961912931
-0.55034841961289 0.667310151837187
0.0270618708595063 0.798570273916677
0.440641653293375 0.523905244197207
0.137142005522338 0.0418905813540118
-0.0710398038977208 -0.186730111328039
-0.188226093054609 0.0427013154610656
-0.131153674188419 0.341747164582727
-0.120140696209008 0.46972881373411
-0.0206773667092816 0.422014998196545
0.0559852575313519 0.152267339033587
-0.0543062892718209 -0.191981015434986
-0.755646813097376 -0.602947426583316
-1.0960806566815 -1.05927870100435
-1.06194014534767 -0.613822867098786
};
\addplot [thick, orangered, opacity=0.7]
table {%
0.130618929862976 -0.883857071399689
0.123250834643841 -1.43454921245575
0.0162579361349344 -1.39750921726227
0.0622384250164032 -0.702845752239227
0.370445668697357 -0.0144643895328045
0.684044301509857 0.412192523479462
0.617954611778259 0.759570598602295
-0.0504062138497829 1.0484721660614
-0.982232093811035 1.19582831859589
-1.57319390773773 1.21838164329529
-1.63391530513763 1.16243231296539
-1.33702206611633 0.979000151157379
-0.893121898174286 0.563976764678955
-0.594079732894897 0.104293182492256
-0.653853118419647 -0.0077935392037034
-0.841062784194946 0.289513677358627
-0.627621054649353 0.756994903087616
0.0499039441347122 0.990022003650665
0.493840992450714 0.64475154876709
0.262860238552094 -0.00143741664942354
-0.152587726712227 -0.300227999687195
-0.22798065841198 -0.0494224242866039
-0.149156451225281 0.394307017326355
-0.133900240063667 0.628220200538635
-0.00820817984640598 0.534865617752075
0.121175937354565 0.191756278276443
-0.156071707606316 -0.318895429372787
-0.825222492218018 -0.850843906402588
-1.31721818447113 -1.04286873340607
-1.27595007419586 -0.639591038227081
};
\addplot [thick, orangered, opacity=0.7, mark=*, mark size=1.5, mark options={solid}]
table {%
0.130618929862976 -0.883857071399689
};
\addplot [thick, orangered, opacity=0.7, mark=triangle*, mark size=1.5, mark options={solid}]
table {%
-1.27595007419586 -0.639591038227081
};
\addplot [thick, dodgerblue, opacity=0.7, mark=*, mark size=1.5, mark options={solid}]
table {%
0.0606423866340622 -0.681447595736459
};
\addplot [thick, dodgerblue, opacity=0.7, mark=triangle*, mark size=1.5, mark options={solid}]
table {%
-1.06194014534767 -0.613822867098786
};
\end{axis}

\end{tikzpicture} & \begin{tikzpicture}

\definecolor{dodgerblue}{RGB}{30,144,255}
\definecolor{orangered}{RGB}{255,69,0}

\begin{axis}[
height=\figureheight,
major tick length=1ex,
tick pos=left,
ticks=none,
width=\figurewidth,
xmin=-2.3, xmax=2.3,
xticklabels={},
ymin=-2.3, ymax=2.3,
yticklabels={}
]
\addplot [thick, dodgerblue, opacity=0.7]
table {%
0.103736373904482 -0.699820184915211
0.160343614471264 -1.31084477818494
0.0542707902691376 -1.22966794332267
0.0369566855110611 -0.533303719355442
0.240334015151918 0.01982465192668
0.485498759676871 0.423264122116477
0.442934957569613 0.571822045918532
-0.130639646066452 0.854810006311202
-0.724122089533472 1.05508178304867
-1.80340219017002 1.26195177599011
-1.95633949172124 1.28296939323719
-1.33300025209577 0.932173696458281
-0.82710835165499 0.491731675076923
-0.49554376269899 0.087507936612091
-0.563715716678774 0.0151141998956519
-0.796236640622402 0.22866789518475
-0.591271548751517 0.628747509252375
-0.0116455946512936 0.81447209427435
0.41708942245432 0.569201558135116
0.0912670162182546 0.00245110589594544
-0.0885975641997563 -0.142459767679237
-0.181155761920149 0.001052192667577
-0.158245386015902 0.270612567705434
-0.0876608712210483 0.500191088926844
-0.0924860061319251 0.420008260788739
0.0566459181018374 0.174254926565584
-0.0694101082651563 -0.175793331556387
-0.787538661664464 -0.744231394375375
-1.21069850629326 -1.00745252265989
-1.16630712800253 -0.592064764739131
};
\addplot [thick, orangered, opacity=0.7]
table {%
0.130618929862976 -0.883857071399689
0.123250834643841 -1.43454921245575
0.0162579361349344 -1.39750921726227
0.0622384250164032 -0.702845752239227
0.370445668697357 -0.0144643895328045
0.684044301509857 0.412192523479462
0.617954611778259 0.759570598602295
-0.0504062138497829 1.0484721660614
-0.982232093811035 1.19582831859589
-1.57319390773773 1.21838164329529
-1.63391530513763 1.16243231296539
-1.33702206611633 0.979000151157379
-0.893121898174286 0.563976764678955
-0.594079732894897 0.104293182492256
-0.653853118419647 -0.0077935392037034
-0.841062784194946 0.289513677358627
-0.627621054649353 0.756994903087616
0.0499039441347122 0.990022003650665
0.493840992450714 0.64475154876709
0.262860238552094 -0.00143741664942354
-0.152587726712227 -0.300227999687195
-0.22798065841198 -0.0494224242866039
-0.149156451225281 0.394307017326355
-0.133900240063667 0.628220200538635
-0.00820817984640598 0.534865617752075
0.121175937354565 0.191756278276443
-0.156071707606316 -0.318895429372787
-0.825222492218018 -0.850843906402588
-1.31721818447113 -1.04286873340607
-1.27595007419586 -0.639591038227081
};
\addplot [thick, orangered, opacity=0.7, mark=*, mark size=1.5, mark options={solid}]
table {%
0.130618929862976 -0.883857071399689
};
\addplot [thick, orangered, opacity=0.7, mark=triangle*, mark size=1.5, mark options={solid}]
table {%
-1.27595007419586 -0.639591038227081
};
\addplot [thick, dodgerblue, opacity=0.7, mark=*, mark size=1.5, mark options={solid}]
table {%
0.103736373904482 -0.699820184915211
};
\addplot [thick, dodgerblue, opacity=0.7, mark=triangle*, mark size=1.5, mark options={solid}]
table {%
-1.16630712800253 -0.592064764739131
};
\end{axis}

\end{tikzpicture} \\ [-12.5pt]

            \raisebox{-7.7pt}{\includegraphics[clip,width=0.179\linewidth]{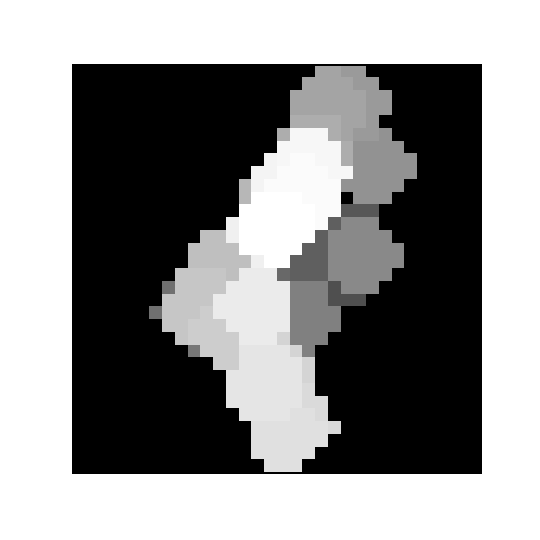}}                                                                                                                                        & \begin{tikzpicture}

\definecolor{dodgerblue}{RGB}{30,144,255}
\definecolor{orangered}{RGB}{255,69,0}

\begin{axis}[
height=\figureheight,
major tick length=1ex,
tick pos=left,
ticks=none,
width=\figurewidth,
xmin=-2.3, xmax=2.3,
xticklabels={},
ymin=-2.3, ymax=2.3,
yticklabels={}
]
\addplot [thick, dodgerblue, opacity=0.7]
table {%
0.576589769057303 -0.0537016272721874
-0.217421319383118 -0.0755701816044567
-0.416735801757975 -0.0724031325986108
0.0500994048101825 0.00641341462263062
0.461443415785378 0.0524620056692289
1.07919115063967 -0.0317099941505628
0.558010621288779 -0.145732605323624
-0.874568303841956 -0.12850558063659
-1.86691205926081 -0.0847440823605386
-0.874873932454743 -0.0822671925611411
0.0631953937782038 0.0213549441083416
1.45504757066856 -0.0402264426245951
1.13318495528434 0.0727021110935055
0.300953567719328 0.132353553863633
0.2591631730302 0.146650529389315
0.639301393997673 0.0327274996886179
0.260984454863629 -0.0470722154694824
-1.04448311489791 -0.0858863176531198
-1.78165420241134 -0.130683849296164
-0.6972596351115 -0.0576948672551658
-0.0581936624658984 0.0303659261634368
0.104400316253907 0.0649934979599577
0.0602960677169554 0.0625274164931717
-0.130440031444751 0.0251845540854961
-0.527219204589741 -0.0619062226300694
-0.055467721594111 -0.114209012276668
0.475382122016882 -0.00560651981465307
0.642733754517618 0.0234147681108151
0.511539020563298 -0.0216521678748857
0.0506447821317689 -0.12496869795979
};
\addplot [thick, orangered, opacity=0.7]
table {%
0.366939178802573 0.544308889075379
-0.0393922363635109 0.153044414766944
-0.171470129722819 -0.255982419179488
0.217139077812657 -0.443866919940082
0.690060662029928 -0.132685621924137
0.766995609035767 0.292622654257457
0.346148451463463 0.172271935533689
-0.45141227990693 -0.280440149836286
-1.09665179220435 -0.581931605387172
-0.835240478545047 -0.770738409949477
0.172176235264808 -0.66929066527765
0.97616350891657 0.0778237018075399
1.1050391453169 1.19562388234451
0.809234789368935 1.92533909442776
0.522440879179164 1.88898553888503
0.336332090647201 1.38086004088896
-0.0794115577628089 0.706119838185789
-0.722278092333986 -0.0151191774516462
-1.00252521038247 -0.551385248261815
-0.653905393045869 -0.870899401900271
-0.0968474272888374 -1.34413719672837
0.165250036289711 -1.97347498714028
0.0413178383714442 -2.11119065172911
-0.220560097375275 -1.48772246805243
-0.332608921775628 -0.540260247350335
-0.170964603032204 0.373292028169541
0.128056016557671 1.02083697110411
0.324992733268306 1.13356075889309
0.23979727825081 0.804233832000487
-0.104156260536759 0.42256844624534
};
\addplot [thick, orangered, opacity=0.7, mark=*, mark size=1.5, mark options={solid}]
table {%
0.366939178802573 0.544308889075379
};
\addplot [thick, orangered, opacity=0.7, mark=triangle*, mark size=1.5, mark options={solid}]
table {%
-0.104156260536759 0.42256844624534
};
\addplot [thick, dodgerblue, opacity=0.7, mark=*, mark size=1.5, mark options={solid}]
table {%
0.576589769057303 -0.0537016272721874
};
\addplot [thick, dodgerblue, opacity=0.7, mark=triangle*, mark size=1.5, mark options={solid}]
table {%
0.0506447821317689 -0.12496869795979
};
\end{axis}

\end{tikzpicture} & \begin{tikzpicture}

\definecolor{dodgerblue}{RGB}{30,144,255}
\definecolor{orangered}{RGB}{255,69,0}

\begin{axis}[
height=\figureheight,
major tick length=1ex,
tick pos=left,
ticks=none,
width=\figurewidth,
xmin=-2.3, xmax=2.3,
xticklabels={},
ymin=-2.3, ymax=2.3,
yticklabels={}
]
\addplot [thick, dodgerblue, opacity=0.7]
table {%
0.18302225646332 0.20944185676385
-0.0421766661465618 0.075502813764463
-0.0732439319359057 -0.112951002996102
0.32318106223275 -0.267767249453656
0.641736974429904 -0.143943528228303
0.598814756049676 0.0561370236085573
0.294061917819097 0.0128823069526501
-0.302605246640349 -0.0742533061419369
-0.834629270079867 -0.0981204174050265
-0.577311882930877 -0.314923714534569
0.309622036895516 -0.449820701124889
0.815563252046331 -0.090674512621547
0.544832832620715 0.385428348458242
0.0656857088420985 0.591839989273164
-0.139705906325501 0.621854102287485
-0.157270884456295 0.530044247037891
-0.23413728217839 0.343685334560431
-0.519301165686311 0.0986119892644094
-0.755793490979596 -0.142653057046185
-0.455535684383345 -0.37060466698828
0.176221899650071 -1.00252383560327
0.500991578882414 -1.94744150261316
0.369762097446641 -2.10017180085546
0.0421667359610663 -1.20733588774786
-0.193948935692674 -0.268591494751876
-0.243289047203324 0.214357901740608
-0.166313133819147 0.422595821745954
-0.031043911870633 0.440055476512128
-0.0174705342217879 0.333171797908236
-0.190148912651205 0.224196400192358
};
\addplot [thick, orangered, opacity=0.7]
table {%
0.366939178802573 0.544308889075379
-0.0393922363635109 0.153044414766944
-0.171470129722819 -0.255982419179488
0.217139077812657 -0.443866919940082
0.690060662029928 -0.132685621924137
0.766995609035767 0.292622654257457
0.346148451463463 0.172271935533689
-0.45141227990693 -0.280440149836286
-1.09665179220435 -0.581931605387172
-0.835240478545047 -0.770738409949477
0.172176235264808 -0.66929066527765
0.97616350891657 0.0778237018075399
1.1050391453169 1.19562388234451
0.809234789368935 1.92533909442776
0.522440879179164 1.88898553888503
0.336332090647201 1.38086004088896
-0.0794115577628089 0.706119838185789
-0.722278092333986 -0.0151191774516462
-1.00252521038247 -0.551385248261815
-0.653905393045869 -0.870899401900271
-0.0968474272888374 -1.34413719672837
0.165250036289711 -1.97347498714028
0.0413178383714442 -2.11119065172911
-0.220560097375275 -1.48772246805243
-0.332608921775628 -0.540260247350335
-0.170964603032204 0.373292028169541
0.128056016557671 1.02083697110411
0.324992733268306 1.13356075889309
0.23979727825081 0.804233832000487
-0.104156260536759 0.42256844624534
};
\addplot [thick, orangered, opacity=0.7, mark=*, mark size=1.5, mark options={solid}]
table {%
0.366939178802573 0.544308889075379
};
\addplot [thick, orangered, opacity=0.7, mark=triangle*, mark size=1.5, mark options={solid}]
table {%
-0.104156260536759 0.42256844624534
};
\addplot [thick, dodgerblue, opacity=0.7, mark=*, mark size=1.5, mark options={solid}]
table {%
0.18302225646332 0.20944185676385
};
\addplot [thick, dodgerblue, opacity=0.7, mark=triangle*, mark size=1.5, mark options={solid}]
table {%
-0.190148912651205 0.224196400192358
};
\end{axis}

\end{tikzpicture} & \begin{tikzpicture}

\definecolor{dodgerblue}{RGB}{30,144,255}
\definecolor{orangered}{RGB}{255,69,0}

\begin{axis}[
height=\figureheight,
major tick length=1ex,
tick pos=left,
ticks=none,
width=\figurewidth,
xmin=-2.3, xmax=2.3,
xticklabels={},
ymin=-2.3, ymax=2.3,
yticklabels={}
]
\addplot [thick, dodgerblue, opacity=0.7]
table {%
0.0259111893564968 0.351542886125121
0.046347365399245 0.38915774714565
0.0432174887550972 0.348679240278807
-0.00898911255583998 0.246404431611694
-0.132032327043979 0.0868476365209397
-0.31833072240001 -0.115228033880893
-0.530489534508822 -0.324958819893629
-0.72372382410206 -0.504025738808134
-0.8348135996077 -0.597896065328528
-0.739924337717322 -0.49835222601837
-0.29965736503877 -0.0850681864823239
0.462178734718714 0.617102999654591
1.24432790997726 1.33366430834251
1.64165955713963 1.69239051958139
1.48155836721592 1.53345320968108
0.940142741456358 1.0201905283616
0.321105002463916 0.431350119324745
-0.195910973422149 -0.0707268223019311
-0.575549817939391 -0.452894533230786
-0.818509420616211 -0.713516333366992
-0.922883654797378 -0.849474526265241
-0.905128338530526 -0.867469963500014
-0.784160620896204 -0.764983763924203
-0.560341056977536 -0.523379664070447
-0.254220574698825 -0.163069890610721
0.0575303931422015 0.217096606699241
0.28687092811317 0.500186799449473
0.40193608553017 0.639564434736896
0.423566965696138 0.654027044921751
0.375804889006268 0.562590942488701
};
\addplot [thick, orangered, opacity=0.7]
table {%
0.366939178802573 0.544308889075379
-0.0393922363635109 0.153044414766944
-0.171470129722819 -0.255982419179488
0.217139077812657 -0.443866919940082
0.690060662029928 -0.132685621924137
0.766995609035767 0.292622654257457
0.346148451463463 0.172271935533689
-0.45141227990693 -0.280440149836286
-1.09665179220435 -0.581931605387172
-0.835240478545047 -0.770738409949477
0.172176235264808 -0.66929066527765
0.97616350891657 0.0778237018075399
1.1050391453169 1.19562388234451
0.809234789368935 1.92533909442776
0.522440879179164 1.88898553888503
0.336332090647201 1.38086004088896
-0.0794115577628089 0.706119838185789
-0.722278092333986 -0.0151191774516462
-1.00252521038247 -0.551385248261815
-0.653905393045869 -0.870899401900271
-0.0968474272888374 -1.34413719672837
0.165250036289711 -1.97347498714028
0.0413178383714442 -2.11119065172911
-0.220560097375275 -1.48772246805243
-0.332608921775628 -0.540260247350335
-0.170964603032204 0.373292028169541
0.128056016557671 1.02083697110411
0.324992733268306 1.13356075889309
0.23979727825081 0.804233832000487
-0.104156260536759 0.42256844624534
};
\addplot [thick, orangered, opacity=0.7, mark=*, mark size=1.5, mark options={solid}]
table {%
0.366939178802573 0.544308889075379
};
\addplot [thick, orangered, opacity=0.7, mark=triangle*, mark size=1.5, mark options={solid}]
table {%
-0.104156260536759 0.42256844624534
};
\addplot [thick, dodgerblue, opacity=0.7, mark=*, mark size=1.5, mark options={solid}]
table {%
0.0259111893564968 0.351542886125121
};
\addplot [thick, dodgerblue, opacity=0.7, mark=triangle*, mark size=1.5, mark options={solid}]
table {%
0.375804889006268 0.562590942488701
};
\end{axis}

\end{tikzpicture} & \begin{tikzpicture}

\definecolor{dodgerblue}{RGB}{30,144,255}
\definecolor{orangered}{RGB}{255,69,0}

\begin{axis}[
height=\figureheight,
major tick length=1ex,
tick pos=left,
ticks=none,
width=\figurewidth,
xmin=-2.3, xmax=2.3,
xticklabels={},
ymin=-2.3, ymax=2.3,
yticklabels={}
]
\addplot [thick, dodgerblue, opacity=0.7]
table {%
0.111206387265996 -0.375872695811582
-0.38767302937472 -0.00857168384383621
-0.185001477290716 -0.176974625503035
-0.0150199835222009 -0.550603891862338
-0.215684871591104 -0.716317418827704
0.164661672512432 0.145816912261305
-0.482335747167641 -0.229134785428593
-0.161985159694122 -0.113745520162789
-0.0774910449968283 0.0855676855025951
0.0814785272240321 -0.14390414859236
0.0869264285462644 -0.674334921934539
0.338145044840497 -0.10827955417506
0.233501025768517 0.590707953605401
-0.127524592589362 1.52624822429706
-0.212267583252151 1.49389337805503
-0.0554536493283899 0.767845906634616
-0.74098335724737 0.370733695725726
-0.275614717018206 0.124000919450246
-0.0182802544170947 0.00138492125177569
0.0417419504603866 -0.144893546663452
0.411981341026284 -0.771456137064332
0.832310971775027 -1.26717730400112
0.835744549172649 -1.19810109444718
0.504202295128087 -0.736612147158395
-0.0539576296503021 -0.237277300874104
-0.504306511728824 0.147829713062302
-0.116065234110138 0.577838555036291
-0.0348492508589271 0.642531419296402
-0.0307801714652598 0.420665127195816
-0.563453370825918 0.160328919514358
};
\addplot [thick, orangered, opacity=0.7]
table {%
0.366939187049866 0.54430890083313
-0.0393922366201878 0.153044417500496
-0.171470135450363 -0.255982428789139
0.217139080166817 -0.443866908550262
0.690060675144196 -0.132685616612434
0.76699560880661 0.292622655630112
0.346148461103439 0.172271937131882
-0.451412290334702 -0.280440151691437
-1.09665179252625 -0.581931591033936
-0.835240483283997 -0.770738422870636
0.172176241874695 -0.669290661811829
0.976163506507874 0.0778236985206604
1.10503911972046 1.19562387466431
0.809234797954559 1.92533910274506
0.522440850734711 1.88898551464081
0.336332082748413 1.38086009025574
-0.0794115588068962 0.70611983537674
-0.722278118133545 -0.0151191772893071
-1.00252521038055 -0.551385223865509
-0.653905391693115 -0.870899379253387
-0.0968474298715591 -1.34413719177246
0.165250033140182 -1.97347497940063
0.0413178391754627 -2.11119055747986
-0.220560103654861 -1.48772251605988
-0.332608908414841 -0.540260255336761
-0.170964598655701 0.373292028903961
0.128056019544601 1.02083694934845
0.324992746114731 1.13356077671051
0.239797279238701 0.804233849048615
-0.104156263172626 0.422568440437317
};
\addplot [thick, orangered, opacity=0.7, mark=*, mark size=1.5, mark options={solid}]
table {%
0.366939187049866 0.54430890083313
};
\addplot [thick, orangered, opacity=0.7, mark=triangle*, mark size=1.5, mark options={solid}]
table {%
-0.104156263172626 0.422568440437317
};
\addplot [thick, dodgerblue, opacity=0.7, mark=*, mark size=1.5, mark options={solid}]
table {%
0.111206387265996 -0.375872695811582
};
\addplot [thick, dodgerblue, opacity=0.7, mark=triangle*, mark size=1.5, mark options={solid}]
table {%
-0.563453370825918 0.160328919514358
};
\end{axis}

\end{tikzpicture} & \begin{tikzpicture}

\definecolor{dodgerblue}{RGB}{30,144,255}
\definecolor{orangered}{RGB}{255,69,0}

\begin{axis}[
height=\figureheight,
major tick length=1ex,
tick pos=left,
ticks=none,
width=\figurewidth,
xmin=-2.3, xmax=2.3,
xticklabels={},
ymin=-2.3, ymax=2.3,
yticklabels={}
]
\addplot [thick, dodgerblue, opacity=0.7]
table {%
0.25213195843224 0.41658577236344
-0.0154822027805899 0.167550142559931
-0.131691998502964 -0.111604026384295
0.11117787371539 -0.279441501581011
0.483051045490315 -0.0953212272686583
0.594189035593924 0.223156485179071
0.211247902675846 0.175940378736177
-0.325982351555833 -0.148370827657371
-0.949528951635009 -0.483132072857928
-0.705998508067162 -0.613418933295707
0.11914737219231 -0.484942685167314
0.787994486257976 0.0166468872022374
1.01010730584176 1.11248750662884
0.897326449725411 2.10127722697959
0.529812753043439 1.93143764334547
0.276602364313964 1.16785088423912
-0.0925346669002229 0.571352679371007
-0.57951421378942 0.0408214677310998
-0.841201716452314 -0.463730842040878
-0.532550284950457 -0.670555457317678
-0.11835573491057 -1.1331608349988
0.0897712448033554 -2.04228368030404
0.0853650320312613 -2.38579563213267
-0.231796299075298 -1.29500955486151
-0.271294216790605 -0.389155169608274
-0.200358799306513 0.306857429211447
0.0394968476671485 0.791853139841907
0.287613678684776 0.904123515687412
0.154428271373451 0.656469841786154
-0.110149204043274 0.334301342909282
};
\addplot [thick, orangered, opacity=0.7]
table {%
0.366939187049866 0.54430890083313
-0.0393922366201878 0.153044417500496
-0.171470135450363 -0.255982428789139
0.217139080166817 -0.443866908550262
0.690060675144196 -0.132685616612434
0.76699560880661 0.292622655630112
0.346148461103439 0.172271937131882
-0.451412290334702 -0.280440151691437
-1.09665179252625 -0.581931591033936
-0.835240483283997 -0.770738422870636
0.172176241874695 -0.669290661811829
0.976163506507874 0.0778236985206604
1.10503911972046 1.19562387466431
0.809234797954559 1.92533910274506
0.522440850734711 1.88898551464081
0.336332082748413 1.38086009025574
-0.0794115588068962 0.70611983537674
-0.722278118133545 -0.0151191772893071
-1.00252521038055 -0.551385223865509
-0.653905391693115 -0.870899379253387
-0.0968474298715591 -1.34413719177246
0.165250033140182 -1.97347497940063
0.0413178391754627 -2.11119055747986
-0.220560103654861 -1.48772251605988
-0.332608908414841 -0.540260255336761
-0.170964598655701 0.373292028903961
0.128056019544601 1.02083694934845
0.324992746114731 1.13356077671051
0.239797279238701 0.804233849048615
-0.104156263172626 0.422568440437317
};
\addplot [thick, orangered, opacity=0.7, mark=*, mark size=1.5, mark options={solid}]
table {%
0.366939187049866 0.54430890083313
};
\addplot [thick, orangered, opacity=0.7, mark=triangle*, mark size=1.5, mark options={solid}]
table {%
-0.104156263172626 0.422568440437317
};
\addplot [thick, dodgerblue, opacity=0.7, mark=*, mark size=1.5, mark options={solid}]
table {%
0.25213195843224 0.41658577236344
};
\addplot [thick, dodgerblue, opacity=0.7, mark=triangle*, mark size=1.5, mark options={solid}]
table {%
-0.110149204043274 0.334301342909282
};
\end{axis}

\end{tikzpicture} & \begin{tikzpicture}

\definecolor{dodgerblue}{RGB}{30,144,255}
\definecolor{orangered}{RGB}{255,69,0}

\begin{axis}[
height=\figureheight,
major tick length=1ex,
tick pos=left,
ticks=none,
width=\figurewidth,
xmin=-2.3, xmax=2.3,
xticklabels={},
ymin=-2.3, ymax=2.3,
yticklabels={}
]
\addplot [thick, dodgerblue, opacity=0.7]
table {%
0.246796225070504 0.446050761594103
-0.0106076071566509 0.146727487698032
-0.156450127647652 -0.122911955993566
0.139585155067367 -0.307200012886485
0.559263203292198 -0.111131883358096
0.567095892864068 0.298746449101334
0.219247945643436 0.124721686437549
-0.362143852218242 -0.166848464259204
-1.00239227135131 -0.459859972775596
-0.744669143242468 -0.670693157116069
0.141247285741345 -0.5426329527436
0.766647009375403 0.0565081923654318
1.05224774672622 0.980225675498868
0.821114641965579 2.07190563932649
0.48838730584038 1.85948668444934
0.282186672105116 1.19381906532584
-0.110329957832628 0.578939142353728
-0.61285929907537 0.00348936725985607
-0.875961765415003 -0.466886086841035
-0.561484293951918 -0.706530789366552
-0.0565674414977043 -1.23620572810465
0.169356757849979 -2.26519674448217
0.0980955213292004 -2.52863484752317
-0.210635637082289 -1.37969337830521
-0.281182768847312 -0.432889676052549
-0.199551936453721 0.295795176991437
0.0411839098372182 0.832692332776483
0.286080738711665 0.977962799125699
0.137685697700935 0.66384986618302
-0.070823795258526 0.307687988419075
};
\addplot [thick, orangered, opacity=0.7]
table {%
0.366939187049866 0.54430890083313
-0.0393922366201878 0.153044417500496
-0.171470135450363 -0.255982428789139
0.217139080166817 -0.443866908550262
0.690060675144196 -0.132685616612434
0.76699560880661 0.292622655630112
0.346148461103439 0.172271937131882
-0.451412290334702 -0.280440151691437
-1.09665179252625 -0.581931591033936
-0.835240483283997 -0.770738422870636
0.172176241874695 -0.669290661811829
0.976163506507874 0.0778236985206604
1.10503911972046 1.19562387466431
0.809234797954559 1.92533910274506
0.522440850734711 1.88898551464081
0.336332082748413 1.38086009025574
-0.0794115588068962 0.70611983537674
-0.722278118133545 -0.0151191772893071
-1.00252521038055 -0.551385223865509
-0.653905391693115 -0.870899379253387
-0.0968474298715591 -1.34413719177246
0.165250033140182 -1.97347497940063
0.0413178391754627 -2.11119055747986
-0.220560103654861 -1.48772251605988
-0.332608908414841 -0.540260255336761
-0.170964598655701 0.373292028903961
0.128056019544601 1.02083694934845
0.324992746114731 1.13356077671051
0.239797279238701 0.804233849048615
-0.104156263172626 0.422568440437317
};
\addplot [thick, orangered, opacity=0.7, mark=*, mark size=1.5, mark options={solid}]
table {%
0.366939187049866 0.54430890083313
};
\addplot [thick, orangered, opacity=0.7, mark=triangle*, mark size=1.5, mark options={solid}]
table {%
-0.104156263172626 0.422568440437317
};
\addplot [thick, dodgerblue, opacity=0.7, mark=*, mark size=1.5, mark options={solid}]
table {%
0.246796225070504 0.446050761594103
};
\addplot [thick, dodgerblue, opacity=0.7, mark=triangle*, mark size=1.5, mark options={solid}]
table {%
-0.070823795258526 0.307687988419075
};
\end{axis}

\end{tikzpicture} \\ [-6.5pt]
            \bottomrule
        \end{tabular}
        }
    \end{center}
    \vspace{-1ex}
}
    \end{minipage}
    \hfill
    \begin{minipage}{0.28\textwidth}
        \tikzexternaldisable
        \centering
        \scriptsize
        \setlength{\figurewidth}{5.2cm}
        \setlength{\figureheight}{3.5cm}
        \tikzexternaldisable
        \includegraphics[width=0.92\textwidth]{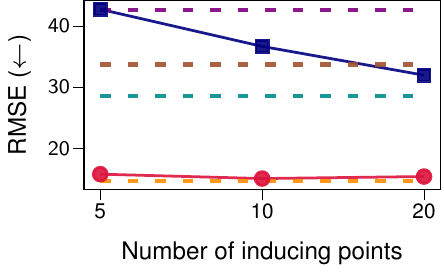}
        \tikzexternalenable
        \vspace{1ex}
        {
            \tiny
            
\definecolor{color_1}{HTML}{df2b4f}
\definecolor{color_2}{HTML}{19198c}
\definecolor{color_3}{HTML}{aa6143}
\definecolor{color_4}{HTML}{900289}
\definecolor{color_5}{HTML}{ff9a36}
\definecolor{color_6}{HTML}{009695}
\tikzexternaldisable
{\setlength{\tabcolsep}{1.8pt}
    \begin{tabular}{clclcl}
        \toprule

        {\protect\tikz[baseline=-1ex]\protect\draw[color=color_vae, fill=color_vae, opacity=0.99, mark size=1.7pt, line width=1.4pt] plot[] (-0.0,0)--(-0.15,0);} & \textsf{VAE} &
        {\protect\tikz[baseline=-1ex]\protect\draw[color=color_gpvae, fill=color_gpvae,  opacity=0.99, mark size=1.7pt, line width=1.4pt] plot[] (-0.0,0)--(-0.15,0);} &  \textsf{GPVAE} &
        {\protect\tikz[baseline=-1ex]\protect\draw[color=color_svgp_vae, fill=color_svgp_vae, mark=square*, opacity=0.99, mark size=1.7pt, line width=0.0pt] plot[] (-0.1,0);} &  \textsf{SVGP-VAE} \\
        {\protect\tikz[baseline=-1ex]\protect\draw[color=color_bae, fill=color_bae, opacity=0.99, mark size=1.7pt, line width=1.4pt] plot[] (-0.0,0)--(-0.15,0);} & \textsf{BAE} &
        {\protect\tikz[baseline=-1ex]\protect\draw[color=color_gpbae, fill=color_gpbae, opacity=0.99, mark size=1.7pt, line width=1.4pt] plot[] (-0.0,0)--(-0.15,0);} & \textsf{GP-BAE} &
        {\protect\tikz[baseline=-1ex]\protect\draw[color=color_bsgpae, fill=color_bsgpae, mark=*, opacity=0.99, mark size=1.7pt, line width=0.0pt] plot[] (-0.1,0);} & \textsf{SGP}-\textsf{BAE} \\
        \bottomrule
    \end{tabular}\tikzexternalenable
}
        }
        \captionof{figure}{Performance of autoencoder models as a function of the number of inducing points.
                    \label{fig:moving_ball_rmse}
        }
    \end{minipage}
    \vspace{-2ex}
\end{figure*}

\vspace{-2ex}

\paragraph{Extension for missing data.}
In practice, real-world data may be sparse, with many missing and few overlapping dimensions across the entire dataset. 
We can easily extend $\gls{SGPBAE}$ to handle such datasets.
We assume that any possible permutation of observed features is potentially missing, such that each high-dimensional observation $\mby_n = \mby^{o}_{n} \cup \mby^{u}_{n}$ contains a set of observed features $\mby^{o}_n$ and unobserved features $\mby^{u}_n$.
The likelihood term of the inference objective (\cref{eq:energy_bsgpae}) factorizes across data points and dimensions, so
there is no major modification in this objective, as the summation of the likelihood term should be done only over the non-missing dimensions, i.e.:
\vspace{-0ex}
\begin{align}
    p(\mbY \g \mbZ, \mbvarphi) = \sum_{n=1}^{N} \log p(\mby_n^o \g \mbz_n, \mbvarphi). 
\end{align}
\vspace{-5ex}

\section{Experiments} \label{sec:experiments}

In this section, we provide empirical evidence that our \gls{SGPBAE} outperforms alternatives of combination between \gls{GP} priors and \gls{AE} models on synthetic data and real-world high-dimensional benchmark datasets.
Throughout all experiments, unless otherwise specified, we use the \gls{RBF} kernel with \gls{ARD} with marginal variance and independent lengthscales $\lambda_i$ per feature \citep{Mackay96}.
We place a lognormal prior with unit variance and means equal to and $0.05$ for the lengthscales and variance, respectively.
Since the auxiliary data of most of the considered datasets  are timestamps, we impose a non-informative uniform prior on the inducing inputs.
We observe that this prior works well in our experiments.
We set the hyperparameters of the number of \gls{SGHMC} and optimizer steps $J=30$, and $K=50$, respectively.
The details for all experiments are available in \cref{sec:experimental_details}.

\vspace{-1ex}

\subsection{Synthetic moving ball data}

We begin our empirical evaluation by considering the moving ball dataset proposed by \citet{Pearce19}.
This dataset comprises grayscale videos showing the movement of a ball.
The two-dimensional trajectory of the ball is simulated from a \gls{GP} characterized by an \gls{RBF} kernel. 
Our task is to reconstruct the underlying trajectory in the 2D latent space from the frames in pixel space.
Unlike \citet{Jazbec21a}, we generate a fixed number of $35$ videos for training and another $35$ videos for testing, rather than generating new training videos at each iteration, regardless of the fact that tens of thousands of iterations are performed. 
This new setting is more realistic and is designed to show the data efficiency of the considered methods.
It is still possible to perform full \gls{GP} inference on such a small dataset.
Therefore, we consider full GP-based methods, such as \gls{GPBAE} and \textsc{gp}-\textsc{vae} \citep{Pearce19}, as oracles for the sparse variants. 
Because the dataset is quite small, we perform full-batch training/inference.

\vspace{-2ex}

\paragraph{Are Bayesian AEs better than variational counterparts?}
In this experiment, we show that, by relaxing strong assumptions on the posterior of latent space and taking advantage of a powerful scalable \gls{MCMC} method, \glspl{BAE} consistently outperform \glspl{VAE}.
\cref{fig:moving_ball_rmse} illustrates the performance of the considered methods in terms of \gls{RMSE}.
The results show that our \gls{GPBAE} model performs much better than  \textsc{gp}-\textsc{vae} \citep{Pearce19} though both models use the same full \gls{GP} priors.
In addition, by treating inducing inputs and kernel hyper-parameters of sparse \glspl{GP} in a Bayesian fashion, \gls{SGPBAE} offers a rich modeling capability.
This is evident in the improved performance of \gls{SGPBAE} compared to sparse variational Gaussian process \gls{VAE} (\svgpvae) \citep{Jazbec21a} and is able to reach closely to the \gls{GPBAE} performance despite using a small number of inducing points.
These numerical results align with the qualitative evaluation of the reconstructed trajectories shown in \cref{tab:moving_ball}.
As expected, the standard \gls{BAE} and \gls{VAE} endowed with a $\cN(\mathbold{0}, \mbI)$ prior on the latent space completely fail to model the trajectories faithfully.
In contrast, \gls{GPBAE} and \gls{SGPBAE} are able to match them closely.

\vspace{-2ex}

\begin{figure}[t]
    \tikzexternaldisable
    \centering
    \scriptsize
    \setlength{\figurewidth}{8.9cm}
    \setlength{\figureheight}{2.9cm}
    \tikzexternaldisable
    \includegraphics[width=0.43\textwidth]{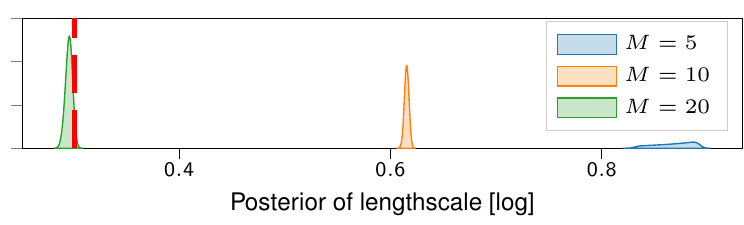}
    \tikzexternalenable
    \vspace{-1ex}
    \caption{The posterior of the lengthscale corresponding to using a different number of inducing points, $M$.
            The \textcolor{red}{\textbf{red}} line denotes the true lengthscale.
            \label{fig:posterior_lengthscale}}
    
    \vspace{-3ex}
\end{figure}

\begin{figure}[t]
    \tikzexternaldisable
    \centering
    \scriptsize
    \setlength{\figurewidth}{8.7cm}
    \setlength{\figureheight}{2.3cm}
    \tikzexternaldisable
    \includegraphics[width=0.42\textwidth]{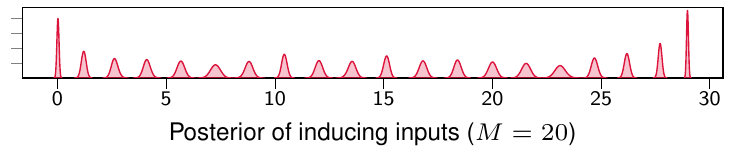}
    \tikzexternalenable
    \vspace{-1ex}
    \caption{The posterior of the inducing position.
            \label{fig:posterior_inducing}}
    \vspace{-5ex}
\end{figure}

\paragraph{Benefits of being fully Bayesian.} 
Our \gls{SGPBAE} model has the same advantage as the \svgpvae \citep{Jazbec21a} in that it allows for an arbitrary \gls{GP} kernel, and the kernel hyperparameters and inducing inputs can be inferred jointly during training or inference. 
In contrast, other methods either use fixed \gls{GP} priors \citep{Pearce19} or employ a two-stage approach \citep{Casale18}, where the \gls{GP} hyperparameters are optimized separately from the \glspl{AE}.
\svgpvae optimizes these hyperparameters using the common practice of maximization of the marginal likelihood, \textsc{ml}-\textsc{ii}.
This results in a point estimate of the hyperparameters but may be prone to overfitting, especially when the training data is small while there are many hyperparameters.
A distinct advantage of \gls{SGPBAE} over \svgpvae is that it is fully Bayesian for the \gls{GP} hyperparameters and inducing points.
This not only improves the quality of the predictions but also offers sensible uncertainty quantification. 
\cref{fig:posterior_lengthscale} illustrates the posterior of the lengthscale.
By using a sufficient number of inducing points, \gls{SGPBAE} is able to recover the true lengthscale.
When using too few inducing points, the model tends to estimate a larger lengthscale.
This is expected as the effective lengthscale of the observed process in the subspace spanned by these few inducing points is larger. 
We also observe that the more inducing points are used, the more confident the posterior is over the lengthscale. 
Our method also produces a sensible posterior distribution on the inducing positions, as shown in \cref{fig:posterior_inducing}.
The estimated inducing positions are evenly spaced over the time dimension, which is reasonable since the latent trajectories are generated from stationary \glspl{GP}.

\vspace{-1.75ex}

\subsection{Conditional generation of rotated  MNIST}
\begin{figure*}
    \begin{minipage}{0.63\textwidth}
        {
    \captionof{table}{Conditionally generated \mnist images. 
    The right most column depicts the epistemic uncertainty obtained by our \gls{SGPBAE} model.
    \label{tab:rotated_mnist}}
    \footnotesize
    \vspace{-2ex}
    \begin{center}
        \scalebox{.78}
        {
        \setlength{\tabcolsep}{-3pt}
        \begin{tabular}{ccccccc}
            \toprule
            \textsc{gt image}                                                                             & \textsc{deep}-\textsc{svigp}                                                                                & \tikzexternaldisable ({\protect\tikz[baseline=-.65ex]\protect\draw[thick, color=color_cvae, fill=color_cvae, mark=*, mark size=2pt, line width=1.25pt] plot[] (-.0, 0)--(-0,0);}) \tikzexternalenable \textsc{cvae}                                                                                 & \tikzexternaldisable ({\protect\tikz[baseline=-.65ex]\protect\draw[thick, color=color_gppvae, fill=color_gppvae, mark=*, mark size=2pt, line width=1.25pt] plot[] (-.0, 0)--(-0,0);}) \tikzexternalenable \textsc{gppvae}                                                                               & \tikzexternaldisable ({\protect\tikz[baseline=-.65ex]\protect\draw[thick, color=color_svgp_vae, fill=color_svgp_vae, mark=*, mark size=2pt, line width=1.25pt] plot[] (-.0, 0)--(-0,0);}) \tikzexternalenable \textsc{svgp}-\textsc{vae}                                                                              & \tikzexternaldisable ({\protect\tikz[baseline=-.65ex]\protect\draw[thick, color=color_bsgpae, fill=color_bsgpae, mark=*, mark size=2pt, line width=1.25pt] plot[] (-.0, 0)--(-0,0);}) \tikzexternalenable \textsc{sgp}-\textsc{bae}                                                                               & \tikzexternaldisable ({\protect\tikz[baseline=-.65ex]\protect\draw[thick, color=color_bsgpae, fill=color_bsgpae, mark=*, mark size=2pt, line width=1.25pt] plot[] (-.0, 0)--(-0,0);}) \tikzexternalenable \textsc{var.}                                                                             \\
            \midrule                                                                                                                                                                                                                                                                                                                                                                                                                                                                                                                                                                                                                                                                                      \\ [-18.5pt]
            \raisebox{0pt}{\includegraphics[clip,width=0.19\linewidth]{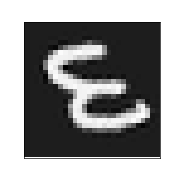}} & \raisebox{0pt}{\includegraphics[clip,width=0.19\linewidth]{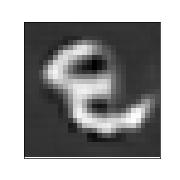}} & \raisebox{0pt}{\includegraphics[clip,width=0.19\linewidth]{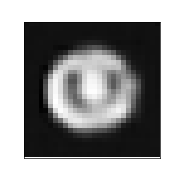}} & \raisebox{0pt}{\includegraphics[clip,width=0.19\linewidth]{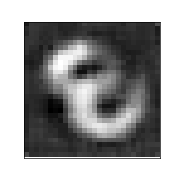}} & \raisebox{0pt}{\includegraphics[clip,width=0.19\linewidth]{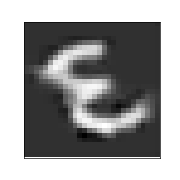}} & \raisebox{0pt}{\includegraphics[clip,width=0.19\linewidth]{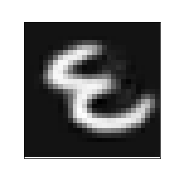}} & \raisebox{0pt}{\includegraphics[clip,width=0.19\linewidth]{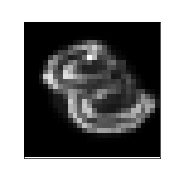}} \\ [-12.5pt]
            \raisebox{0pt}{\includegraphics[clip,width=0.19\linewidth]{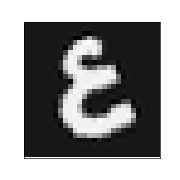}} & \raisebox{0pt}{\includegraphics[clip,width=0.19\linewidth]{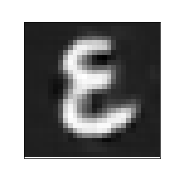}} & \raisebox{0pt}{\includegraphics[clip,width=0.19\linewidth]{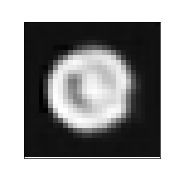}} & \raisebox{0pt}{\includegraphics[clip,width=0.19\linewidth]{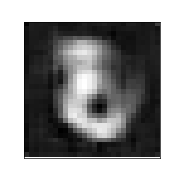}} & \raisebox{0pt}{\includegraphics[clip,width=0.19\linewidth]{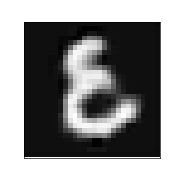}} & \raisebox{0pt}{\includegraphics[clip,width=0.19\linewidth]{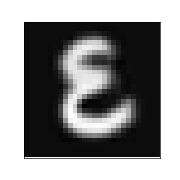}} & \raisebox{0pt}{\includegraphics[clip,width=0.19\linewidth]{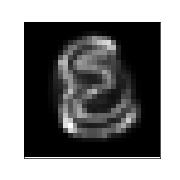}} \\ [-12.5pt]
            \raisebox{0pt}{\includegraphics[clip,width=0.19\linewidth]{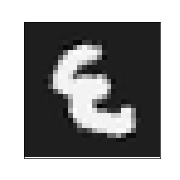}} & \raisebox{0pt}{\includegraphics[clip,width=0.19\linewidth]{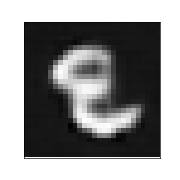}} & \raisebox{0pt}{\includegraphics[clip,width=0.19\linewidth]{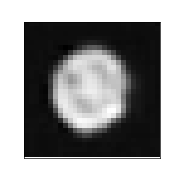}} & \raisebox{0pt}{\includegraphics[clip,width=0.19\linewidth]{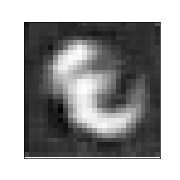}} & \raisebox{0pt}{\includegraphics[clip,width=0.19\linewidth]{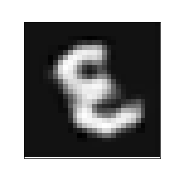}} & \raisebox{0pt}{\includegraphics[clip,width=0.19\linewidth]{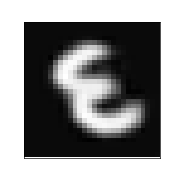}} & \raisebox{0pt}{\includegraphics[clip,width=0.19\linewidth]{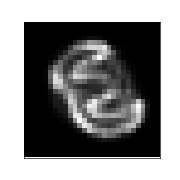}} \\ [-10.5pt]
            \bottomrule
        \end{tabular}
        }
    \end{center}
}

    \end{minipage}
    \hfill
    \begin{minipage}{0.34\textwidth}
       \tikzexternaldisable
        \centering
        \scriptsize
        \setlength{\figurewidth}{6.1cm}
        \setlength{\figureheight}{3.8cm}
        \tikzexternaldisable
        \includegraphics[width=0.99\textwidth]{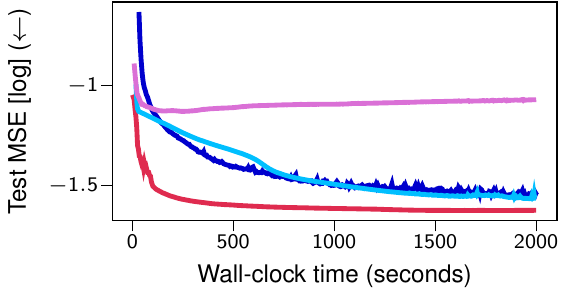}
        \tikzexternalenable
        {
            \tiny
            \definecolor{color_svigp}{HTML}{00bfff}

\tikzexternaldisable
{\setlength{\tabcolsep}{1.8pt}
    \begin{tabular}{clcl}
        \toprule
        {\protect\tikz[baseline=-1ex]\protect\draw[color=color_cvae, fill=color_cvae, opacity=0.99, mark size=1.7pt, line width=2.1pt] plot[] (-0.0,0)--(-0.15,0);} & \textsf{CVAE} &
        {\protect\tikz[baseline=-1ex]\protect\draw[color=color_svgp_vae, fill=color_svgp_vae, opacity=0.99, mark size=1.7pt, line width=2.1pt] plot[] (-0.0,0)--(-0.15,0);} & \textsf{SVGP-VAE} \\
        {\protect\tikz[baseline=-1ex]\protect\draw[color=color_svigp, fill=color_svigp, opacity=0.99, mark size=1.7pt, line width=2.1pt] plot[] (-0.0,0)--(-0.15,0);} & \textsf{DEEP-SVIGP} &
        {\protect\tikz[baseline=-1ex]\protect\draw[color=color_bsgpae, fill=color_bsgpae, opacity=0.99, mark size=1.7pt, line width=2.1pt] plot[] (-0.0,0)--(-0.15,0);} & \textsf{SGP-BAE \textbf{(Ours)}} \\
        \bottomrule
    \end{tabular}\tikzexternalenable
}
        }
        \captionof{figure}{Comparison of test \gls{MSE} on the rotated \mnist dataset as function of training time. \label{fig:mnist_convergence}}
        
    \end{minipage}
    \vspace{-1ex}
\end{figure*}

In the next experiment, we consider a large-scale benchmark of conditional generation.
We follow the experimental setup from \citep{Casale18, Jazbec21a}, where they use a rotated \mnist dataset ($N=4050$). \. %
In particular, we are given a set of images of digit three that have been rotated at different angles in $[0, 2\pi)$.
Our goal is to generate a new image rotated at an unseen angle.
As it is not trivial to apply full \gls{GP} for such a large dataset, we omit \gpvae \citep{Pearce19} and \gls{GPBAE}  baselines.
We consider the \gppvae model \cite{Casale18}, which employs a low-rank approximation for the \gls{GP}, and the \svgpvae model \citep{Jazbec21a} as baselines.
For both \svgpvae and \gls{SGPBAE}, we use a number of inducing points of $M = 32$ and a mini-batch size of $B = 256$.
We also compare to \gls{CVAE} \citep{Sohn15}, which allows conditional generation tasks.
Following \citet{Jazbec21a}, we consider an extension of the sparse \gls{GP} model \citep{Hensman13}, named  \deepsvigp, that utilizes a deep likelihood parameterized by a neural network. %
As shown in \cref{tab:rotated_mnist}, our \gls{SGPBAE} model generates images that are more visually appealing and most faithful to the ground truth when compared to other approaches.

\begin{table}
    \caption{Results on the rotated \mnist digit 3 dataset.
        Here, we report the mean and standard deviation computed from 4 runs.
        \label{tab:rotated_mnist_mse}
    }
    \vspace{-2ex}
    \small
    \begin{center}
        \scalebox{.85}{
        \rowcolors{1}{}{mylightgray}
        \begin{tabular}{rlc}
            \toprule
                                                                                                                                                                                                                & \textsc{model}                               & \textsc{mse}  ($\downarrow$)                \\
            \midrule
            \midrule
            \tikzexternaldisable ({\protect\tikz[baseline=-.65ex]\protect\draw[thick, color=color_cvae, fill=color_cvae, mark=*, mark size=2pt, line width=1.25pt] plot[] (-.0, 0)--(-0,0);}) \tikzexternalenable & \textsc{cvae} \citep{Sohn15}                 & 0.0819 $\pm$ 0.0027          \\
            \tikzexternaldisable ({\protect\tikz[baseline=-.65ex]\protect\draw[thick, color=color_gppvae, fill=color_gppvae, mark=*, mark size=2pt, line width=1.25pt] plot[] (-.0, 0)--(-0,0);}) \tikzexternalenable & \textsc{gppvae} \citep{Casale18}              & 0.0351 $\pm$ 0.0005          \\
            \tikzexternaldisable ({\protect\tikz[baseline=-.65ex]\protect\draw[thick, color=color_svgp_vae, fill=color_svgp_vae, mark=*, mark size=2pt, line width=1.25pt] plot[] (-.0, 0)--(-0,0);}) \tikzexternalenable & \textsc{svgp}-\textsc{vae} \citep{Jazbec21a} & 0.0257 $\pm$ 0.0004          \\
            \tikzexternaldisable ({\protect\tikz[baseline=-.65ex]\protect\draw[thick, color=color_bsgpae, fill=color_bsgpae, mark=*, mark size=2pt, line width=1.25pt] plot[] (-.0, 0)--(-0,0);}) \tikzexternalenable & \textsc{sgp-bae \small{\textbf{(ours)}}}                       & \textbf{0.0228 $\pm$ 0.0004} \\
            \midrule
                                                                                                                                                                                                                & \textsc{deep}-\textsc{svigp} \citep{Jazbec21a}             & 0.0236 $\pm$ 0.0010          \\

            \bottomrule
        \end{tabular}
        }
    \end{center}
    \vspace{-4.5ex}
\end{table}

\begin{table*}[t]
    \caption{A comparison between methods of multi-output \gls{GP} models and \gls{GP} autoencoders on the \textsc{eeg} and \textsc{jura} datasets.
    \label{tab:eeg_and_jura}}
    \vspace{-2ex}
    \small
    \begin{center}
        \scalebox{.90}
        {
        \begin{tabular}{cr||cc|ccc}
            \toprule
            \textsc{dataset}               & \textsc{metric}             & \textsc{igp}         & \textsc{gpar}            & \tikzexternaldisable ({\protect\tikz[baseline=-.65ex]\protect\draw[thick, color=color_sgp_vae, fill=color_sgp_vae, mark=*, mark size=2pt, line width=1.25pt] plot[] (-.0, 0)--(-0,0);}) \tikzexternalenable \textsc{sgp}-\textsc{vae} & \tikzexternaldisable ({\protect\tikz[baseline=-.65ex]\protect\draw[thick, color=color_bsgpae, fill=color_bsgpae, mark=*, mark size=2pt, line width=1.25pt] plot[] (-.0, 0)--(-0,0);}) \tikzexternalenable \textsc{sgp-bae \scriptsize{(\textbf{ours})}} & \textsc{dsgp-bae \scriptsize{(\textbf{ours})}} \\
            \midrule
            \midrule
            \multirow{2}{*}{\textsc{eeg}}  &
            \textsc{smse} ($\downarrow$)   & 1.70 $\pm$ 0.14             & 0.28 $\pm$ 0.00      & 0.52 $\pm$ 0.05          & \textbf{0.22 $\pm$ 0.01} & \textbf{0.21 $\pm$ 0.01}                                                                                                                                                                                                                                                                                                                                                                                                                                                 \\
                                           & \textsc{nll} ($\downarrow$) & 2.59 $\pm$ 0.02      & \textbf{1.68 $\pm$ 0.01} & 1.98 $\pm$ 0.02                                                                                                                                                                                                                     & 1.96 $\pm$ 0.08                     & 2.25 $\pm$ 0.13                                                                                                                                                                                             \\
            \midrule
            \multirow{2}{*}{\textsc{jura}} & \textsc{mae} ($\downarrow$) & 0.57 $\pm$ 0.00      & \textbf{0.42 $\pm$ 0.01} & 0.54 $\pm$ 0.01                                                                                                                                                                                                                       & \textbf{0.45 $\pm$ 0.03}        & \textbf{0.44 $\pm$ 0.02}                                                                                                                                                                                                 \\
                                           & \textsc{nll} ($\downarrow$) & 1563.42 $\pm$ 166.55 & 17.81 $\pm$ 1.06         & 1.02 $\pm$ 0.01                                                                                                                                                                                                                       & \textbf{0.91 $\pm$ 0.04}        & \textbf{0.85 $\pm$ 0.04}                                                                                                                                                                                                 \\
            \bottomrule
        \end{tabular}
        }
    \end{center}

    \vspace{-4.05ex}
\end{table*}

\vspace{-2ex}
\paragraph{Performance of conditional generation.}
\cref{tab:rotated_mnist_mse} presents the quantitative evaluation of the conditionally generated images in terms of \gls{MSE}.
Our \gls{SGPBAE} model clearly outperforms the other competing methods.
It is worth noting that \deepsvigp \citep{Hensman13} does not use an amortization mechanism, and its performance is considered to be an upper bound for that of \svgpvae. 
As discussed by \citet{Jazbec21a}, \deepsvigp can be used for conditional generation tasks, where the goal is to impose a single \gls{GP} over the entire dataset, and therefore amortization is not necessary. 
However, this model cannot be used in tasks where inference has to be amortized across multiple \glspl{GP}, such as learning interpretable data representations.

\vspace{-2ex}

\paragraph{Computational efficiency.}
Similarly to the competing methods that use sparse approximations such as \svgpvae and \deepsvigp, each iteration of \gls{GPBAE} involves the computation of the inverse covariance matrix, resulting in a time complexity of $\mathcal{O}(M^3)$.
\cref{fig:mnist_convergence} shows the convergence in terms of test \gls{MSE} for the competing methods and our \gls{SGPBAE}, trained for a fixed training time budget.
We omit \gppvae \citep{Casale18} from this comparison, as reported by \citet{Jazbec21a}, which is significantly slower than the sparse methods. 
This result demonstrates that \gls{SGPBAE} converges remarkably faster in terms of wall-clock time while achieving better final predictive performance.

\vspace{-2ex}

\paragraph{Epistemic uncertainty quantification.}
Unlike the competing methods, our \gls{SGPBAE} model can capture epistemic uncertainty by treating the decoder's parameters in a Bayesian manner. 
This capability can be helpful in out-of-distribution settings or when uncertainty quantification on reconstructed or generated images is desired. 
As shown in the rightmost column of \cref{tab:rotated_mnist}, \gls{SGPBAE} can provide sensible epistemic uncertainty quantification.
Our model exhibits increased uncertainty for semantically and visually challenging pixels, such as the boundaries of the digits. 

\vspace{-1ex}

\subsection{Missing data imputation}
Next, we consider the task of imputing missing data on multi-output spatio-temporal datasets.
We compare our method against Sparse \textsc{gp}-\textsc{vae} (\sgpvae) \citep{Ashman20} and multi-output \gls{GP} methods such as Independent \glspl{GP} (\igp) and Gaussian Process Autoregressive Regression (\gpar) \citep{Requeima19}.
Additionally, we consider the \deepsgpbae model, which is an extension of our \gls{SGPBAE} model endowed with $3$-layer \gls{GP} priors.
For a fair comparison, we treat partially missing data as zeros to feed into the inference network (encoder) for \sgpvae and our \gls{SGPBAE} model.
We leave the adoption of partial inference networks \citep{Ashman20} to our model for future work.
We follow the experimental setup of \citet{Requeima19} and \citet{Ashman20} and use two standard datasets for this comparison.

\vspace{-2.2ex}

\paragraph{Electroencephalogram (EEG).}
This dataset comprises $256$ measurements taken over one second.
Each measurement consists of seven electrodes, FZ and F1-F6, placed on the patient’s scalp ($\mbx_n \in \mathbb{R}^{1}$, $\mby_n \in \mathbb{R}^{7}$).
The goal is to predict the last $100$ samples for electrodes FZ, F1, and F2, given that the first $156$ samples of FZ, F1, and F2 and the whole signals of F3-F6 are observed.
Performance is measured with the standardized mean squared error (\smse) and negative log-likelihood (\nll).

\vspace{-2.2ex}

\paragraph{Jura.}
This is a geospatial dataset consisting of $359$ measurements of the topsoil concentrations of three metals --- Nickel, Zinc, and Cadmium --- collected from a region of Swiss Jura ($\mbx_n \in \mathbb{R}^2$, $\mby_n \in \mathbb{R}^3$).
The dataset is split into a training set comprised of Nickel and Zinc measurements for all $359$ locations and Cadmium measurements for just $259$ locations. 
The task is to predict the Cadmium measurements at the remaining $100$ locations conditioned on the observed training data.
Performance is evaluated with the mean absolute error (\mae) and negative log-likelihood.

\cref{tab:eeg_and_jura} compares the performance of \gls{SGPBAE} to the competing methods.
As mentioned in \citet{Ashman20}, \gpar is the state-of-the-art method for these datasets.
We find that our \gls{SGPBAE} and \deepsgpbae  models perform comparably with \gpar on the \textsc{eeg} dataset but better on the \textsc{jura} dataset.
A significant advantage of \gls{GP} autoencoder methods is that they model $P$ outputs using only $C$ latent \glspl{GP}, while \gpar uses $P$ \glspl{GP}. 
This can be beneficial when the dimensionality of the data, $P$, is very high. 
Similarly to the previous experiments, \gls{SGPBAE} consistently performs better than variational approximation-based methods such as \sgpvae \citep{Ashman20}.
As expected, \igp is the worst-performing method due to its inability to model the correlations between output variables.

\vspace{-1.2ex}

\section{Conclusions}

\vspace{-0.5ex}

We have introduced our novel \gls{SGPBAE} that integrates fully Bayesian sparse \glspl{GP} on the latent space of Bayesian autoencoders.
Our proposed model is generic, as it allows an arbitrary \gls{GP} kernel and deep \glspl{GP}.
The inference for this model is carried out by a powerful and scalable \gls{SGHMC} sampler.
Through a rigorous experimental campaign, we have demonstrated the excellent performance of \gls{SGPBAE} on a wide range of representation learning and generative modeling tasks.

\vspace{-2.75ex}

\paragraph{Limitations and future works.}
While our model's ability to learn disentangled representations has been demonstrated through empirical evidence, there is a need to establish a theoretical guarantee for the disentanglement of its latent space.
Furthermore, it would be useful to study the amortization gap \citep{Cremer2018, Krishnan18a,marino2018iterative} of our model.
Additionally, exploring more informative priors for the decoder's parameters \citep{Tran21}, beyond the isotropic Gaussian prior used in this work, is also worthwhile.
As a further direction for future research, it would be interesting to apply the model to downstream applications where modeling correlations between data points and uncertainty quantification are required, such as in bioinformatics and climate modeling.

\section*{Acknowledgements}
We thank Simone Rossi and Dimitrios Milios for helpful discussions.
SM acknowledges support by the National Science Foundation (NSF) under an NSF CAREER Award, award numbers 2003237 and 2007719, by the Department of Energy under grant DE-SC0022331, by the HPI Research Center in Machine Learning and Data Science at UC Irvine, and by gifts from Qualcomm and Disney.
MF gratefully acknowledges support from the AXA Research Fund and the Agence Nationale de la Recherche (grant ANR-18-CE46-0002 and ANR-19-P3IA-0002).

\bibliography{main}
\bibliographystyle{icml2023}

\newpage
\appendix
\onecolumn
\glsresetall
\section{A taxonomy of latent variable models}

\begin{figure}[h]
    \centering
    \includegraphics[width=0.98\textwidth]{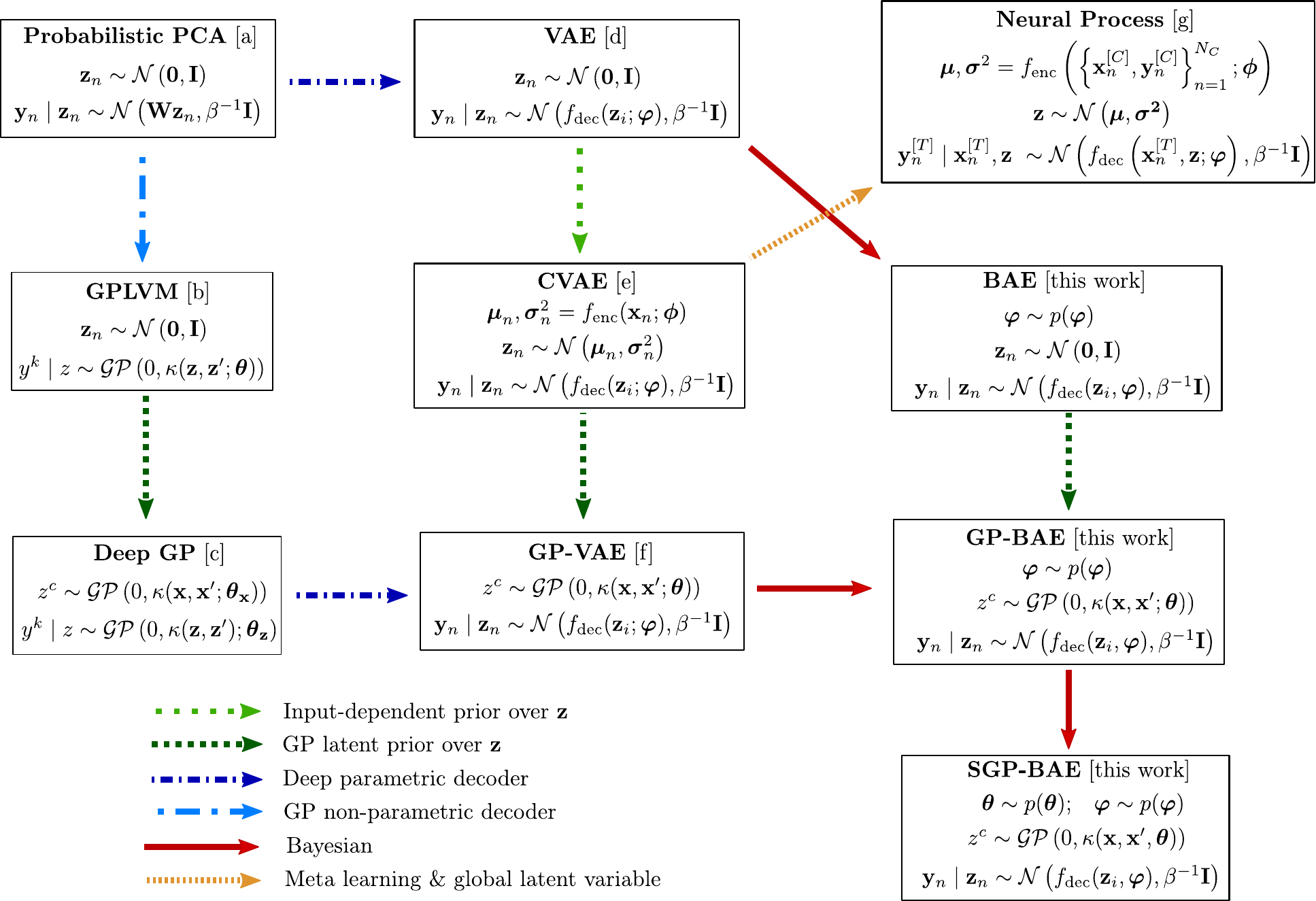}
    \caption{Connections between our proposed models and other latent variables models.  References are [a] for \citet{bishop2006pattern}, [b] for \citet{Lawrence05}, [c] for \citet{Damianou13}, [d] for \citet{Kingma14, RezendeMW14}, [e] for \citet{Sohn15}, [f] for \citet{Casale18,Pearce19,Jazbec21a}, and [g] for \citet{garnelo2018neural}.
    \label{fig:taxonomy_lvms}}
\end{figure}

By considering the characteristics of the prior distribution on latent variables, the likelihood function, input dependencies, and Bayesian treatments,  we can draw connections between our proposed models and other latent variable models.
\cref{fig:taxonomy_lvms} summarizes these relationships.
Here, we assume an isotropic Gaussian likelihood with precision $\beta$ for the high-dimensional observed data $\mby_n$ as used in our experiments.

Probabilistic \gls{PCA} \citep{bishop2006pattern} is a simple latent variable model that imposes an isotropic Gaussian prior over the latent space and linear projection from the latent variables to the observed data.
\glspl{VAE} \citep{Kingma14, RezendeMW14}  build upon this by introducing a non-linear parametric mapping to the observed data, while \glspl{GPLVM} utilize a non-parametric \gls{GP} mapping.
\gls{CVAE} \citep{Sohn15}  is an extension of  \glspl{VAE} that utilizes an input-dependent prior in the latent space for conditional generation tasks. 
However, this model does not account for correlations between latent representations.
Gaussian Process \glspl{VAE} \citep{Casale18,Pearce19,Jazbec21a} overcome this problem by imposing \gls{GP} priors on the latent space.
Our model, \gls{SGPBAE},  further enhances the modeling capabilities of these models by using a fully Bayesian approach for the latent variables, decoder, and \gls{GP} hyper-parameters in a fully Bayesian manner, and decoupling the model from the variational inference formulation.

Latent neural processes \citep{garnelo2018neural} can be seen as an extension of \gls{CVAE}.
However, this model follows a meta-learning approach and splits the data into a context set, $\left\{\mbx_n^{[C]}, \mby_n^{[C]}\right\}_{n=1}^{N_{C}}$, and a target set, $\left\{\mbx_n^{[T]}, \mby_n^{[T]}\right\}_{n=1}^{N_{T}}$.
This model uses a global latent variable $\mbz$ to capture the global properties of the context data.
The likelihood is conditioned on both new target input $\mbx_n^{[T]}$ and the global latent variable $\mbz$.

\label{sec:sghmc}
\section{Details on stochastic gradient Hamiltonian Monte Carlo} 

\gls{HMC} \cite{Neal2011} is a highly-efficient Markov Chain Monte Carlo (MCMC) method used to generate samples $\mbtheta$ from a potential energy function $U(\mbtheta)$.
\gls{HMC} introduces auxiliary momentum variables $\mbr$ and samples are then generated from the joint distribution $p(\mbtheta, \mbr)$ using the Hamiltonian dynamics:
\begin{align}
    \begin{cases}
        d\mbtheta & = \mbM^{-1} \mbr dt,  \\
        d\mbr & = -\nabla U(\mbtheta) dt,
    \end{cases}
\end{align}
where $\mbM$ is an arbitrary mass matrix that serves as a preconditioner.
The continuous system described by \gls{HMC} is approximated using $\eta$-discretized numerical integration, and subsequent Metropolis steps are applied to account for errors that may result from this approximation.

However, \gls{HMC} becomes computationally inefficient when applied to large datasets, as it requires the calculation of the gradient $\nabla U(\mbtheta)$ on the entire dataset.
To address this issue, \cite{Cheni2014} proposed an extension of \gls{HMC} called \gls{SGHMC}, which uses a noisy but unbiased estimate of the gradient $\nabla \tilde{U}(\mbtheta)$ computed from a mini-batch of the data.
The discretized Hamiltonian dynamics are then updated using this estimate of the gradient as follows:
\begin{align}
    \begin{cases}
        \Delta \mbtheta & =      \mbM^{-1} \mbr ,                                                                                         \\
        \Delta \mbr & = - \eta \nabla \tilde{U}(\mbtheta) - \eta \mbC \mbM^{-1} \mbr + \cN(0, 2\eta (\mbC - \tilde{\mbB})),
    \end{cases}
    \label{eq:discretized_hamilton_dynamics}
\end{align}
where $\eta$ is the step size, $\mbC$ is a user-defined friction matrix, $\tilde{\mbB}$ is the estimate for the noise of the gradient evaluation.
To select suitable values for these hyperparameters, we use a scale-adapted version of \gls{SGHMC} proposed by \citet{Springenberg2016},  in which the hyperparameters are automatically adjusted during a burn-in phase. 
After this period, the values of the hyperparameters are fixed.

\paragraph{Estimating $\mbM$.} In our implementation of \gls{SGHMC}, we set the mass matrix $\mbM^{-1} = \mathrm{diag}\left(\hat{V}_{\mbtheta}^{-1/2} \right)$, where $\hat{V}_{\mbtheta}$ is an estimate of the uncentered variance of the gradient, $\hat{V}_{\mbtheta} \approx \mathbb{E}[( \nabla \tilde{U}(\mbtheta))^2] $.
 which can be estimated by using exponential moving average as follows:
\begin{equation}
    \Delta \hat{V}_{\mbtheta} = -\tau^{-1} \hat{V}_{\mbtheta} + \tau^{-1} \nabla (\tilde{U}(\mbtheta))^2,
\end{equation}
where $\tau$ is a parameter vector that specifies the moving average windows.
This parameter can be chosen adaptively using the method proposed by \citet{Springenberg2016} as follows:
\begin{equation}
    \Delta \tau = -g_{\mbtheta}^2 \hat{V}^{-1}_{\mbtheta} \tau + 1, \quad \text{and}, \quad \Delta g_{\mbtheta} = -\tau^{-1} g_{\mbtheta} + \tau^{-1} \nabla \tilde{U}(\mbtheta),
\end{equation}
where $g_{\mbtheta}$ is a smoothed estimate of the gradient $\nabla U(\mbtheta)$.

\paragraph{Estimating $\tilde{\mbB}$.}
Ideally, the estimate of the noise of the gradient evaluation, $\tilde{\mbB}$, should be the empirical Fisher information matrix of $U(\mbtheta)$
However, this matrix is computationally expensive to calculate, so we instead use a diagonal approximation given by $\tilde{\mbB} = \frac{1}{2} \eta \hat{V}_{\mbtheta}$.
This approximation is already obtained from the step of estimating $\mbM$.

\paragraph{Choosing $\mbC$.} In practice, it is common to set the friction matrix $\mbC$ to be equal to the identity matrix, i.e. $\mbC = C \mbI$, which means that the same independent noise is applied to each element of $\mbtheta$.

\paragraph{The discretized Hamiltonian dynamics.} By substituting $\mbv := \eta \hat{V}_{\mbtheta}^{-1/2} \mbr$, the dynamics \cref{eq:discretized_hamilton_dynamics} become
\begin{align}
    \begin{cases}
        \Delta \mbtheta & = \mbv,                                                                                                                                                                           \\
        \Delta \mbv & = -\eta^{2} \hat{V}_{\mbtheta}^{-1/2} \nabla \tilde{U}(\mbtheta) - \eta C \hat{V}_{\mbtheta}^{-1/2}  \mbv + \cN(0, 2 \eta^3 C \hat{V}_{\mbtheta}^{-1} - \eta^4 \mbI).
    \end{cases}
\end{align}
Following \citep{Springenberg2016}, we choose $C$ such that $\eta C \hat{V}_{\mbtheta}^{-1/2} = \alpha \mbI$. This is equivalent to using a constant momentum coefficient of $\alpha$. The final discretized dynamics are then
\begin{align}
    \begin{cases}
        \Delta \mbtheta & = \mbv,                                                                                                                                                     \\
        \Delta \mbv & = -\eta^{2} \hat{V}_{\mbtheta}^{-1/2} \nabla \tilde{U}(\mbtheta) - \alpha  \mbv + \cN(0, 2 \eta^2 \alpha \hat{V}_{\mbtheta}^{-1/2} - \eta^4 \mbI).
    \end{cases}
\end{align}

\section{Details of the scalable sampling objective for sparse Gaussian processes}
\glspl{GP} \citep{Rasmussen06} are one of the main workhorses of Bayesian nonparametric statistics and machine learning, as they provide a flexible and powerful tool for imposing a prior distribution over functions:
\begin{align}
    f(\mbx) \sim \mathcal{GP}(m(\mbx), \kappa(\mbx, \mbx^\prime; \mbtheta)),
\end{align}
where $f(\mbx):\bbR^D\rightarrow\bbR$ maps $D$-dimensional inputs into one-dimensional outputs.
\glspl{GP} are fully characterized by their mean and their covariance:
\begin{align}
    \mathbb{E}[f(\mbx)] = m(\mbx), \quad \quad \text{cov}[f(\mbx), f(\mbx^\prime)] = \kappa(\mbx, \mbx^\prime; \mbtheta),
\end{align}
where $m: \bbR^D\rightarrow\bbR$ is the mean and $k: \bbR^D \times \bbR^D \rightarrow\bbR$ is the kernel function with hyperparameters $\mbtheta$.
\glspl{GP} indeed can be viewed as an extension of a multivariate Gaussian distribution  to infinitely many dimensions, where any fixed set of inputs $\mbX \in \bbR^{N \times D}$ follows the Gaussian distribution:
\begin{align}
    p(\mbf) = \cN(\mathbold{0}, \mbK_{\mbx \mbx \g \mbtheta}).
\end{align}
Here, we assume a zero-mean \gls{GP} prior; $\mbf \in \bbR^{N}$ are the realizations \gls{GP} random variables at the inputs $\mbX$; and $\mbK_{\mbX \mbX \g \mbtheta}$ is the covariance matrix obtained by evaluating $\kappa(\mbx, \mbx^\prime; \mbtheta)$ over all input pairs $\mbx_i$, $\mbx_j$ (we will drop  the explicit parameterization on $\mbtheta$ for the sake of notation).
From now on, in order to keep the notation uncluttered, we focus on a single channel of latent space of \glspl{AE}.
We assume that the latent codes $\mbZ$ are stochastic realizations based on $\mbf$ and additive Gaussian noise i.e. $\mbZ \g \mbf \sim \cN(\mbf, \sigma^2\mbI)$.
We further assume a full factorization $p(\mbZ \g \mbf; \sigma^2 \mbI) = \prod_{n=1}^{N} p(\mbz_n \g f_n; \sigma^2)$.

Though \glspl{GP} provide an elegant mechanism to handle uncertainties, their computational complexity grows cubically with the number of inputs, i.e. $\mathcal{O}(N^3)$.
This problem is commonly tackled by sparse \glspl{GP}, which are a family of approximate models that address the scalability issue by introducing a set of $M$ inducing variables $\mbu = (u_1, ..., u_M)^{\top} \in \bbR^{M\times1}$ at the corresponding inducing inputs $\mbS = [\mbs_1^{\top}, ..., \mbs_M^{\top}]^{\top} \in \bbR^{M\times D}$ with $u_m \equiv f(\mbs_m)$.
The inducing points $\mbS$ can be interpreted as a compressed version of the training data where the number of inducing points $M$ acts as a trade-off parameter between the goodness of the approximation and scalability.
The inducing variables are assumed to be drawn from the same \gls{GP} as the original process, i.e.:
\begin{align}
    p(\mbf, \mbu) &= p(\mbu) p(\mbf \g \mbu), \text{with} \\
    p(\mbu) &= \cN(\mathbold{0}, \Kss), \\ 
    p(\mbf \g \mbu) &= \cN(\Kxs \Kssinv \mbu, \Kxx - \Kxs \Kssinv \Ksx), 
\end{align}
where the covariance matrices $\Kss, \Kxs$ are computed between the elements in $\mbS$ and $\{\mbX, \mbS\}$, respectively, and $\Ksx = \Ksx^{\top}$.

There is a line of works using variational techniques for sparse \glspl{GP} priors for \glspl{VAE} \citep{Jazbec21a, Ashman20}.
More specifically, they rely on the variational framework proposed by \citet{Titsias09a, Hensman13}, enabling the use of \gls{GP} priors on very large datasets.
However, the posterior of the inducing variables $\mbu$  under this framework involves constraining to have a parametric form (usually a Gaussian).

In this work, we take a different route as we treat the inducing variables $\mbu$, inducing inputs $\mbS$ as well as the kernel hyperparameters $\mbtheta$ in a fully Bayesian way.
As discussed in the main paper,  we wish to infer these variables together with the decoders’ parameters and the latent codes by using a powerful and scalable \gls{SGHMC} \citep{Cheni2014} sampler.
To do so, the sampling objective \cref{eq:bsgpae_sampling_obj} should be decomposed over all data points.
The main obstacle is the joint distribution $p(\mbZ, \mbu \g \mbtheta) = \mathbb{E}_{p(\mbf \g \mbu)}[p(\mbZ \g \mbf; \sigma^2 \mbI)]$, which has a complicated form due to the expectation under the conditional $p(\mbf \g \mbu)$.
As discussed by \citet{Rossi21}, this problem can be solved effectively by the fully independent training conditionals \citep[\textsc{fitc}; see][]{quinonero-candela05a}, i.e., parameterizing $p(\mbf \g \mbu)$ as follows:
\begin{align}
    p(\mbf \g \mbu) & \approx \cN(\mbf ; \Kxs \Kssinv \mbu, \text{diag}[\Kxx - \Kxs \Kssinv \Ksx]) \\
                    & = \prod_{n=1}^{N} p(f_n \g \mbu) = \prod_{n=1}^{N} \cN(f_n; \tilde{\mu}_n, \tilde{\sigma}^2_n),
\end{align}
where
\begin{align}
    \tilde{\mu}_n &= \kappa(\mbx_n, \mbS) \Kssinv \mbu, \label{eq:approx_mean} \\ 
    \tilde{\sigma}^2_n &= \kappa(\mbx_n, \mbx_n) - \kappa(\mbx_n, \mbS) \Kssinv \kappa(\mbS, \mbx_n). \label{eq:approx_var}
\end{align}
We then can decompose the log of joint distribution over all data points as follows:
\begin{align}
    \log p(\mbZ, \mbu \g \mbtheta) &= \log \mathbb{E}_{p(\mbf \g \mbu, \mbtheta)}[p(\mbz \g \mbf; \sigma^2\mbI)] \\
    &= \log \int p(\mbf \g \mbu, \mbtheta) p(\mbZ \g \mbf, \sigma^2\mbI) d \mbf \\
    &= \log \int \prod_{n=1}^{N} p(\mbz_n \g f_n; \sigma^2) p(f_n \g \mbu, \mbtheta) df_1...df_n \\
    &= \log \prod_{n=1}^{N} \int \cN(\mbz_n; f_n, \sigma^2) \cN(f_n; \tilde{\mu}_n, \tilde{\sigma}_n^2) df_n \\
    &= \log \prod_{n=1}^{N} \cN(\mbz_n; \tilde{\mu}_n, \tilde{\sigma}_n^2 + \sigma^2) \\
    &= \sum_{n=1}^{N} \log \cN(\mbz_n; \tilde{\mu}_n, \tilde{\sigma}_n^2 + \sigma^2),
\end{align}
where $\tilde{\mu}_n, \tilde{\sigma}_n^2$ are given by \cref{eq:approx_mean} and \cref{eq:approx_var}, respectively.
\label{sec:deep_gp_appendix}
\section{Details of the extension to deep Gaussian processes} 

We assume a deep Gaussian process prior  $f^{(L)} \circ f^{(L-1)} \circ \cdots \circ f^{(1)}$ \citep{Damianou13}, where each $f^{(l)}$ is a \gls{GP}.
For the sake of notation, we use $\mbPsi^{(l)}$ to indicate the set of kernel hyper-parameters and inducing inputs of the $l$-th layer and $\mbf^{(0)}$ as the input $\mbX$.
We additionally denote $\mbPsi = \{ \mbvarphi \} \cup \left\{ \mbu^{(l)}, \mbPsi^{(l)}\right\}_{l=1}^{L}$, where $\mbvarphi$ is the decoder's parameters.

The full joint distribution is as follows:
\vspace{-1ex}
\begin{align}
    p\left(\mbPsi, \{\mbf^{(l)}\}_{l=1}^{L}, \mbZ, \mbY \given \mbX\right) = p(\mbPsi) \underbrace{\prod_{l=1}^{L}  p\left(\mbf^{(l)} \g \mbf^{(l-1)}, \mbPsi\right)  p\left(\mbZ \given \mbf^{(L)}; \sigma^2 \mbI\right)}_{\text{Deep GP prior }} p(\mbY \g \mbZ, \mbvarphi). \label{eq:deep_gp_obj}
\end{align}
Here we omit the dependency on $\mbX$ for notational brevity.

We wish to infer the posterior over $\mbPsi$ and the latent variables $\mbZ$.
To do this, the hidden layers $\mbf^{(l)}$ have to be marginalized and propagated up to the final layer $L$ \citep{Salimbeni17}.
In particular, the log posterior is as follows:
\begin{align}
    \log p(\mbPsi, \mbZ \g \mbY, \mbX)  = \log p(\mbPsi) + \log \E_{p\left(\left\{\mbf^{(l)}\right\}_{l=1}^{L}  \given \left\{\mbu^{(l)}, \mbPsi^{(l)}\right\}_{l=1}^{L}\right)} p\left(\mbZ\given\mbf^{(L)}; \sigma^2\mbI \right)  +  \log p(\mbY \g \mbZ, \mbvarphi) - \log C,
\end{align}
where $C$ is a normalizing constant.

The above posterior is intractable, but we have obtained the form of its (un-normalized) log joint, from which we can sample using the \gls{HMC} method.
Unfortunately, the expectation term is intractable, but we can obtain its estimates using Monte Carlo sampling as follows:
\begin{align}
    & \log \E_{p\left(\left\{\mbf^{(l)}\right\}_{l=1}^{L}  \given \left\{\mbu^{(l)}, \mbPsi^{(l)}\right\}_{l=1}^{L}\right)} p\left(\mbZ\given\mbf^{(L)}; \sigma^2\mbI \right)  \approx \nonumber\\
    & \qquad \approx \log\E_{p\left(\left\{\mbf^{(l)}\right\}_{l=2}^{L}\given\widetilde{\mbf}^{(1)}, \left\{\mbu^{(l)}, \mbPsi^{(l)}\right\}_{l=2}^L\right)}  p\left(\mbZ\given\mbf^{(L)}; \sigma^2\mbI\right) , \qquad \widetilde{\mbf}^{(1)}\sim p\left(\mbf^{(1)}\given\mbu^{(1)}, \mbPsi^{(1)}, \mbf^{(0)}\right), \\
    &  \qquad \approx \log\E_{p\left(\left\{\mbf^{(l)}\right\}_{l=3}^{L}\given\widetilde{\mbf}^{(2)},\left\{\mbu^{(l)}, \mbPsi^{(l)}\right\}_{l=3}^L\right)}  p\left(\mbZ\given\mbf^{(L)}; \sigma^2\mbI\right) , \quad \widetilde{\mbf}^{(2)}\sim p\left(\mbf^{(2)}\given\mbu^{(2)}, \mbPsi^{(2)}, \widetilde{\mbf}^{(1)}\right),\\
    & \qquad \approx \hdots \nonumber \\
    & \qquad \approx \log\E_{p\left(\mbf^{(L)}\given\widetilde{\mbf}^{(L-1)}, \mbu^{(L)}, \mbPsi^{(L)}\right)} p\left(\mbZ\given\mbf^{(L)}; \sigma^2\mbI\right),\quad \widetilde{\mbf}^{(L-1)} \sim p\left(\mbf^{L-1}\left\vert\mbu^{(L-1)}, \mbPsi^{(L-1)}, \widetilde{\mbf}^{(L-2)}\right.\right),\\
    & \qquad = \sum_{n=1}^N \log\E_{p\left(f_n^L\given\widetilde{f}_n^{L-1}, \mbu^{(L)}, \mbPsi^{(L)}\right)} p\left(\mbz_n\given f_n^{(L)}; \sigma^2\right)
\end{align}
Notice that each step of the approximation is unbiased due to the layer-wise factorization of the joint distribution (\cref{eq:deep_gp_obj}).
We can obtain a closed form of the last-layer expectation as $\mbz_n \g f_n^{(L)}$ is a Gaussian.
In the case of using a different distribution, this expectation can be approximated using quadrature \citep{HensmanMG15}.

\label{sec:experimental_details}
\section{Experimental details}

In this section, we present details on implementation and hyperparameters used in our experimental campaign.
All experiments were conducted on a server equipped with a Tesla T4 GPU having $16$ GB RAM.
Throughout all experiments, unless otherwise stated, we impose an isotropic Gaussian prior over the decoder's parameters.
We use the \gls{RBF} kernel with \gls{ARD} with marginal variance and independent lengthscales $\lambda_i$ per feature \citep{Mackay96}.
We place a lognormal prior with unit variance and means equal to and $0.05$ for the lengthscales and variance, respectively.
Since the auxiliary data of most of the considered datasets  are timestamps, we impose a non-informative uniform prior on the inducing inputs.
We observe that this prior works well in our experiments.
The lengthscales are initialized to $1$, whereas the inducing points are initialized by a $k$-means algorithm as commonly used in practice \citep{HensmanMG15}.
For inference, we use an adaptive version of \gls{SGHMC} \citep{Springenberg2016} in which the hyper-parameters are automatically tuned during a burn-in phase.
The hyper-parameters are tuned according to the performance on the validation set.

The random seed for the stochastic inference network is drawn from an isotropic Gaussian distribution, i.e. $q(\mathbold{\varepsilon}) = \cN(\mathbold{0}, \mbI)$.
In case the encoder is a \gls{MLP}, we concatenate the random seeds and the input into a long vector.
The dimension of the random seeds is the same as that of the input.
If the encoder is a convolutional neural network, we spatially stack the random seeds and the input. 
We use an Adam optimizer \citep{jlb2015adam} with a learning rate of $0.001$ for optimizing the encoder network.
We set the hyperparameters of the number of \gls{SGHMC} and encoder optimization steps $J=30$, and $K=50$, respectively.

\subsection{Moving ball experiment}
We use the same network architectures as in \citet{Jazbec21a,Pearce19}.
We follow the data generation procedure of \citet{Jazbec21a}, in which a  squared-exponential \gls{GP} kernel with a lengthscale $l=2$ was used.
Notice that, unlike \citet{Jazbec21a}, we generate a fixed number of $35$ videos for training and another $35$ videos for testing rather than generating new training videos at each iteration, regardless of the fact that tens of thousands of iterations are performed. 
This new setting is more realistic and is designed to show the data efficiency of the considered methods.
The number of frames in each video is $30$.
The dimension of each frame is $32\times 32$.
\cref{tab:moving_ball_settings} presents the default hyperparameters used for our \gls{SGPBAE} and \gls{GPBAE} models.
For the competing methods, we follow the setups specified in \citet{Jazbec21a}.

\subsection{Rotated MNIST experiment}
For the rotated MNIST experiment, we follow the same setup as in \citet{Jazbec21a} and \citet{Casale18}.
In particular, we use the \gls{GP} kernel proposed by \citet{Casale18} and a neural network consisting of three convolutional layers followed by a fully connected layer in the encoder and vice-versa in the decoder.
The details of hyperparameters used for our models are presented in \cref{tab:mnist_settings}.
For the competing methods, we refer to \citet{Jazbec21a}.

\subsection{Missing imputation experiment}
We follow the experimental setting in \citet{Ashman20}, in which squared exponential \gls{GP} kernels are used. 
Notice that, to ensure a fair comparison, we handle partially missing data by treating it as zero and feeding it into the inference network (encoder) for \sgpvae \citep{Ashman20} and our \gls{SGPBAE} model.
This is because it is not trivial to adapt partial inference networks \citep{Ashman20} to our stochastic inference network, and we leave this for future work.
\cref{tab:jura_settings} and \cref{tab:eeg_settings} show the default hyperparameters used for our models.
For multi-output \gls{GP} baselines, we refer to \citet{Requeima19}.

\begin{figure*}[h]
    \begin{minipage}{0.49\textwidth}
        {
    \captionof{table}{Parameter settings for the moving ball experiment.
    \label{tab:moving_ball_settings}}
    \vspace{-2ex}
    \small
    \begin{center}
        \scalebox{.95}
        {
            \rowcolors{1}{}{mylightgray}
            \begin{sc}
                \begin{tabular}{lc}
                    \toprule
                    parameter & value \\                
                    \midrule
                    \midrule
                    num. of feedforward layers in encoder   & $2$ \\
                    num. of feedforward layers in decoder   & $2$ \\
                    width of a hidden feedforward layer   & $500$ \\
                    dimensionality of latent space & $2$ \\
                    activation function & tanh \\
                    \midrule
                    num. of inducing points & $10$ \\
                    mini-batch size & full \\
                    step size & $0.005$ \\
                    momentum & $0.05$ \\
                    num. of burn-in steps & $1500$ \\
                    num. of samples & $100$ \\
                    thinning interval & $400$ \\
                    \bottomrule
                    
                \end{tabular}
            \end{sc}
        }
    \end{center}

    \vspace{-3ex}
}

    \end{minipage}
    \hfill
    \begin{minipage}{0.49\textwidth}
        {
    \captionof{table}{Parameter settings for the rotated \mnist experiment.
    \label{tab:mnist_settings}}
    \vspace{-2ex}
    \small
    \begin{center}
        \scalebox{.95}
        {
            \rowcolors{1}{}{mylightgray}
            \begin{sc}
                \begin{tabular}{lc}
                    \toprule
                    parameter & value \\                
                    \midrule
                    \midrule
                    num. of conv. layers in encoder   & $3$ \\
                    num. of conv. layers in decoder   & $3$ \\
                    num. of filters per conv. channel   & $8$ \\
                    filter size & $3\times 3$ \\
                    num. of feedforward layers in encoder   & $1$ \\
                    num. of feedforward layers in decoder   & $1$ \\
                    activation function & elu \\
                    dimensionality of latent space & $16$ \\
                    \midrule
                    num. of inducing points & $32$ \\
                    mini-batch size & $512$ \\
                    step size & $0.01$ \\
                    momentum & $0.01$    \\
                    num. of burn-in steps & $1500$ \\
                    num. of samples & $200$ \\
                    thinning interval & $800$ \\
                    \bottomrule
                    
                \end{tabular}
            \end{sc}
        }
    \end{center}
}

    \end{minipage}
\end{figure*}

\begin{figure*}[h]
    \begin{minipage}{0.49\textwidth}
            {
        \captionof{table}{Parameter settings for the \textsc{jura} experiment.
        \label{tab:jura_settings}}
        \vspace{-2ex}
        \small
        \begin{center}
            \scalebox{.95}
            {
                \rowcolors{1}{}{mylightgray}
                \begin{sc}
                    \begin{tabular}{lc}
                        \toprule
                        parameter & value \\                
                        \midrule
                        \midrule
                        num. of feedforward layers in encoder   & $1$ \\
                        num. of feedforward layers in decoder   & $2$ \\
                        width of a hidden encoder layer   & $20$ \\
                        width of a hidden decoder layer   & $5$ \\
                        dimensionality of latent space & $2$ \\
                        activation function & relu \\
                        \midrule
                        num. of inducing points & $128$ \\
                        mini-batch size & $100$ \\
                        step size & $0.003$ \\
                        momentum & $0.05$ \\
                        num. of burn-in steps & $1500$ \\
                        num. of samples & $50$ \\
                        thinning interval & $180$ \\
                        \bottomrule
                        
                    \end{tabular}
                \end{sc}
            }
        \end{center}
    
        \vspace{-3ex}
    }

    \end{minipage}
    \hfill
    \begin{minipage}{0.49\textwidth}
            {
        \captionof{table}{Parameter settings for the \textsc{eeg} experiment.
        \label{tab:eeg_settings}}
        \vspace{-2ex}
        \small
        \begin{center}
            \scalebox{.95}
            {
                \rowcolors{1}{}{mylightgray}
                \begin{sc}
                    \begin{tabular}{lc}
                        \toprule
                        parameter & value \\                
                        \midrule
                        \midrule
                        num. of feedforward layers in encoder   & $1$ \\
                        num. of feedforward layers in decoder   & $2$ \\
                        width of a hidden encoder layer   & $20$ \\
                        width of a hidden decoder layer   & $5$ \\
                        dimensionality of latent space & $3$ \\
                        activation function & relu \\
                        \midrule
                        num. of inducing points & $128$ \\
                        mini-batch size & $100$ \\
                        step size & $0.003$ \\
                        momentum & $0.05$ \\
                        num. of burn-in steps & $1500$ \\
                        num. of samples & $50$ \\
                        thinning interval & $180$ \\
                        \bottomrule
                        
                    \end{tabular}
                \end{sc}
            }
        \end{center}
    
        \vspace{-3ex}
    }

    \end{minipage}
\end{figure*}

\section{Additional results}

\subsection{Latent space visualization}
\begin{figure}[t]
    \tikzexternaldisable
    \centering
    \scriptsize
    \setlength{\figurewidth}{13.5cm}
    \setlength{\figureheight}{8.7cm}
    \tikzexternaldisable
    \includegraphics[width=0.75\textwidth]{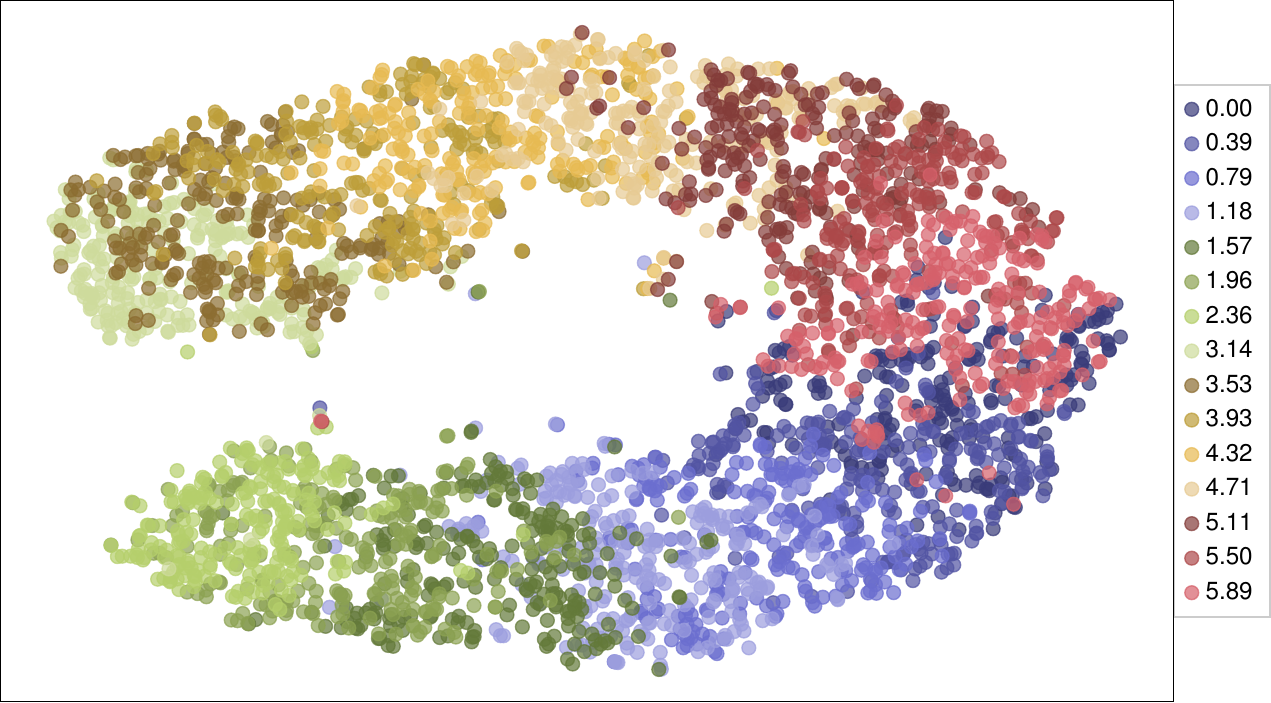}
    \tikzexternalenable
    
    \caption{Visualization of t-SNE embeddings \citep{vandermaaten08a} of \gls{SGPBAE} latent vectors on the training data of the rotated \mnist.
        Each image embedding is colored with respect to its associated angle.
        Here, we use a perplexity parameter of $50$ for t-SNE.
    \label{fig:mnist_tsne}}
\end{figure}

\cref{fig:mnist_tsne} illustrates two-dimensional t-SNE \citep{vandermaaten08a} visualization of latent vectors ($C=16$) for the rotated \mnist data obtained by our \gls{SGPBAE} model.
It is evident that the clusters of embeddings are organized in a clear and orderly manner according to the angles they represent.
Specifically, the embeddings of rotated images are arranged in a continuous sequence from $0$ to $2\pi$ in a clockwise direction.

\subsection{Convergence of SGHMC}
\begin{figure}[t]
    \centering
    \scriptsize
    \setlength{\figurewidth}{6.9cm}
    \setlength{\figureheight}{3.1cm}
    \tikzexternaldisable
    \includegraphics[width=0.34\textwidth]{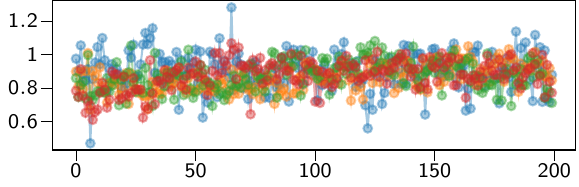}
    \tikzexternalenable
    \setlength{\figurewidth}{3.0cm}
    \setlength{\figureheight}{3.1cm}
    \tikzexternaldisable
    \includegraphics[width=0.109\textwidth]{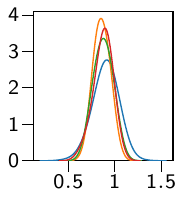}
    \tikzexternalenable
    \centering
    \scriptsize
    \setlength{\figurewidth}{6.9cm}
    \setlength{\figureheight}{3.1cm}
    \tikzexternaldisable
    \includegraphics[width=0.34\textwidth]{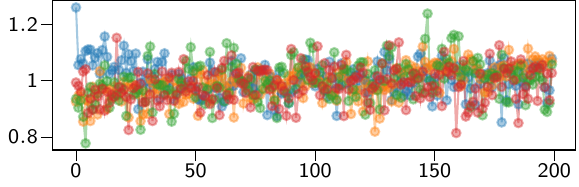}
    \tikzexternalenable
    \setlength{\figurewidth}{3.0cm}
    \setlength{\figureheight}{3.1cm}
    \tikzexternaldisable
    \includegraphics[width=0.111\textwidth]{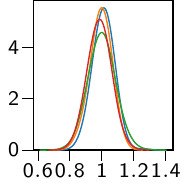}
    \tikzexternalenable

    \vspace{1ex}

    \hspace{-3.0ex}
    \centering
    \scriptsize
    \setlength{\figurewidth}{6.9cm}
    \setlength{\figureheight}{3.1cm}
    \tikzexternaldisable
    \includegraphics[width=0.342\textwidth]{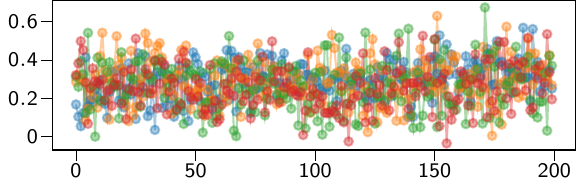}
    \tikzexternalenable
    \setlength{\figurewidth}{3.0cm}
    \setlength{\figureheight}{3.1cm}
    \tikzexternaldisable
    \includegraphics[width=0.108\textwidth]{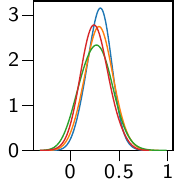}
    \tikzexternalenable
    \tikzexternaldisable
    \centering
    \scriptsize
    \setlength{\figurewidth}{6.9cm}
    \setlength{\figureheight}{3.1cm}
    \tikzexternaldisable
    \includegraphics[width=0.34\textwidth]{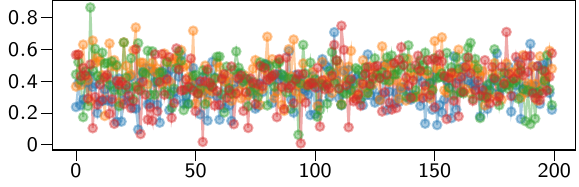}
    \tikzexternalenable
    \setlength{\figurewidth}{3.0cm}
    \setlength{\figureheight}{3.1cm}
    \tikzexternaldisable
    \includegraphics[width=0.102\textwidth]{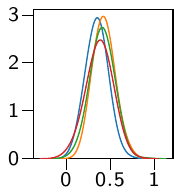}
    \tikzexternalenable
    \caption{Trace plots for four test points on the rotated \mnist dataset. Here, we simulate $4$ chains with $200$ samples for each.  \label{fig:mnist_trace_plot}}
\end{figure}

We follow the common practice using $\hat{R}$ potential scale-reduction diagnostic \citep{Gelman92} to assess the convergence of \gls{MCMC} on \glspl{BNN} \cite{izmailov21a, Tran22}.
Given two or more chains, $\hat{R}$ estimates the ratio between the chain variance  and the average within-chain variance.
For the large-scale \mnist experiment, we compute the $\hat{R}$ statistics on the predictive posterior over four independent chains and obtain a median value of less than $1.1$, which suggests good convergence \cite{izmailov21a}.
In addition, we present traceplots of the predictive posterior in \cref{fig:mnist_trace_plot}, which also support the conclusion of good mixing.

\subsection{Visualization of EEG data}

\begin{figure}[t]
    \begin{subfigure}{\textwidth}
        \tikzexternaldisable
        \centering
        \scriptsize
        \setlength{\figurewidth}{17.5cm}
        \setlength{\figureheight}{3.2cm}
        \tikzexternaldisable
        \includegraphics[width=0.99\textwidth]{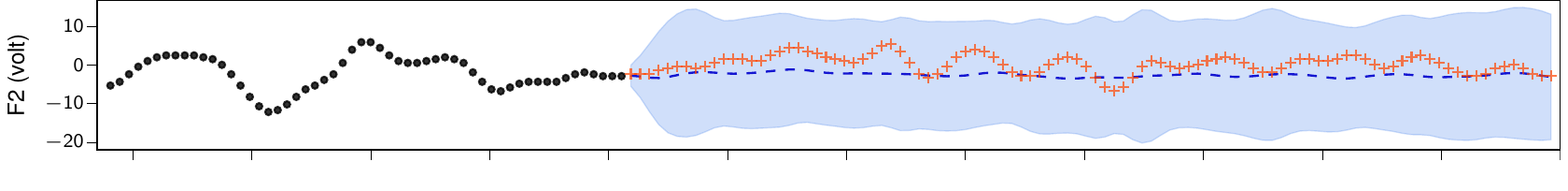}

        \vspace{-2.8ex}
        
        \includegraphics[width=0.99\textwidth]{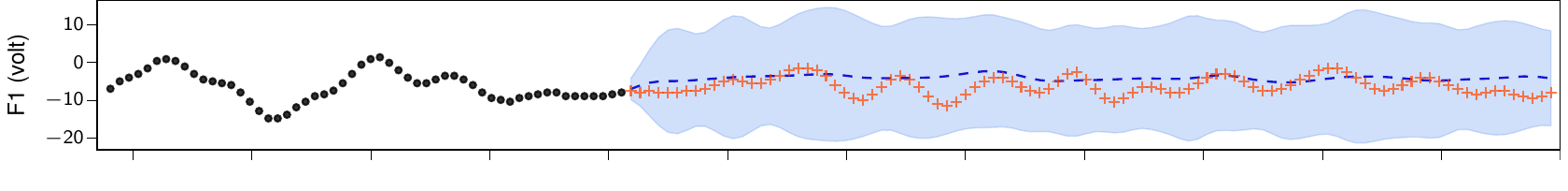}

        \vspace{-3ex}
        
        \includegraphics[width=0.994\textwidth]{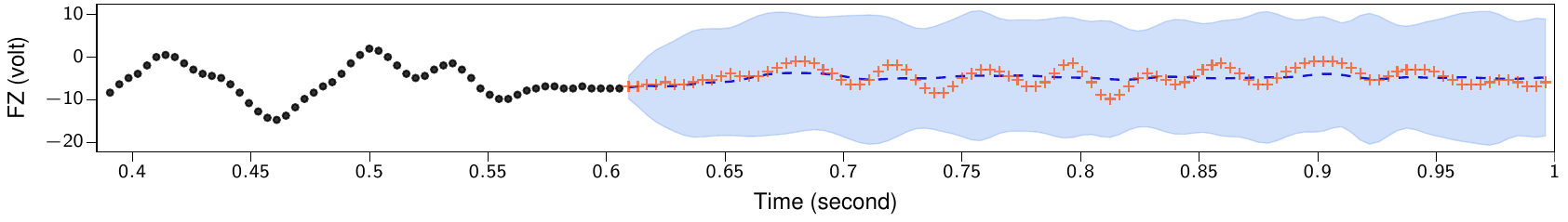}
        \tikzexternalenable
        
        \caption{Independent Gaussian Processes (\textsc{igp})}
    \end{subfigure}

    \vspace{3ex}

    \begin{subfigure}{\textwidth}
        \tikzexternaldisable
        \centering
        \scriptsize
        \setlength{\figurewidth}{17.5cm}
        \setlength{\figureheight}{3.2cm}
        \tikzexternaldisable
        \includegraphics[width=0.99\textwidth]{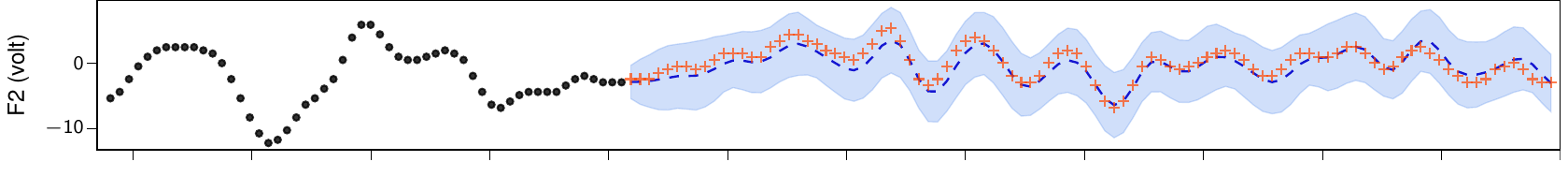}

        \vspace{-4.2ex}
        
        \includegraphics[width=0.99\textwidth]{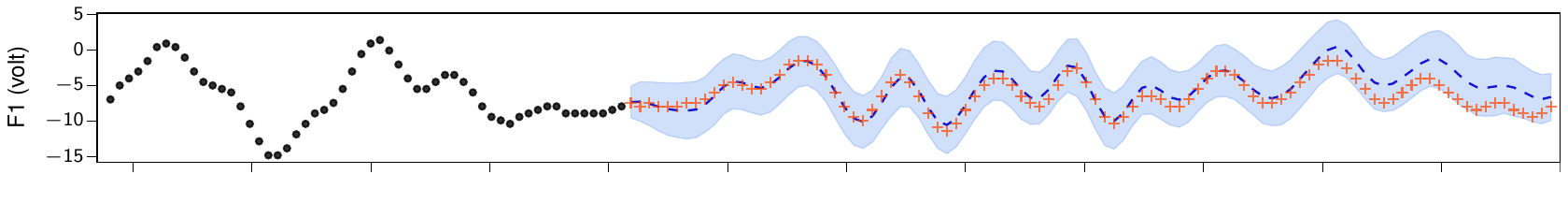}

        \vspace{-3ex}
        
        \includegraphics[width=0.999\textwidth]{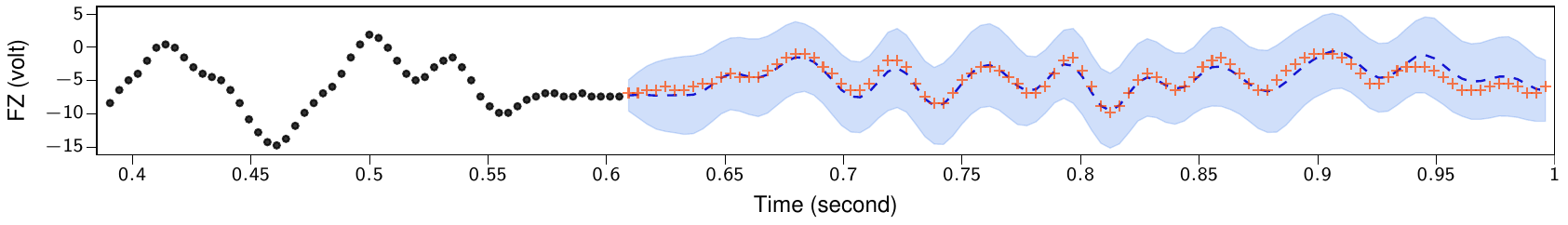}
        \tikzexternalenable

        \caption{Gaussian Process Autoregressive Model (\textsc{gpar})}
    \end{subfigure}

    \vspace{3ex}

    \begin{subfigure}{\textwidth}
        \tikzexternaldisable
        \centering
        \scriptsize
        \setlength{\figurewidth}{17.5cm}
        \setlength{\figureheight}{3.2cm}
        \tikzexternaldisable
        \includegraphics[width=0.99\textwidth]{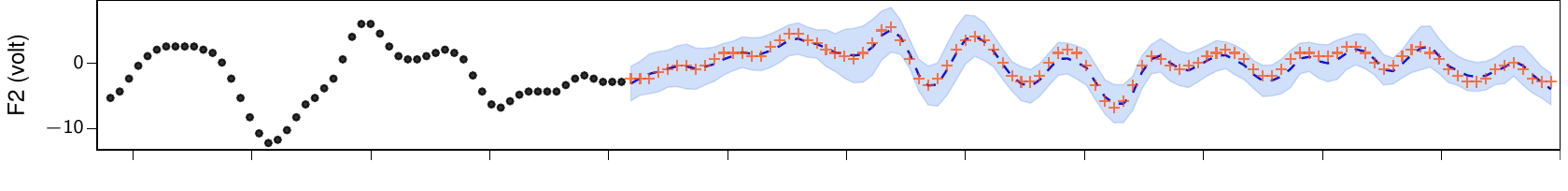}

        \vspace{-4.2ex}
        
        \includegraphics[width=0.99\textwidth]{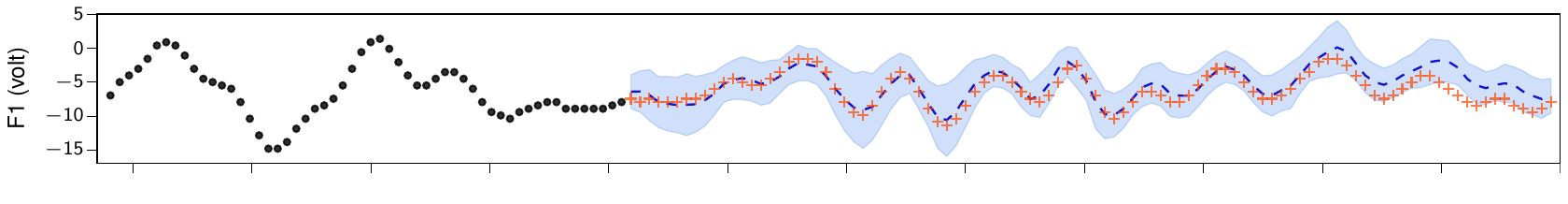}

        \vspace{-3ex}
        
        \includegraphics[width=0.999\textwidth]{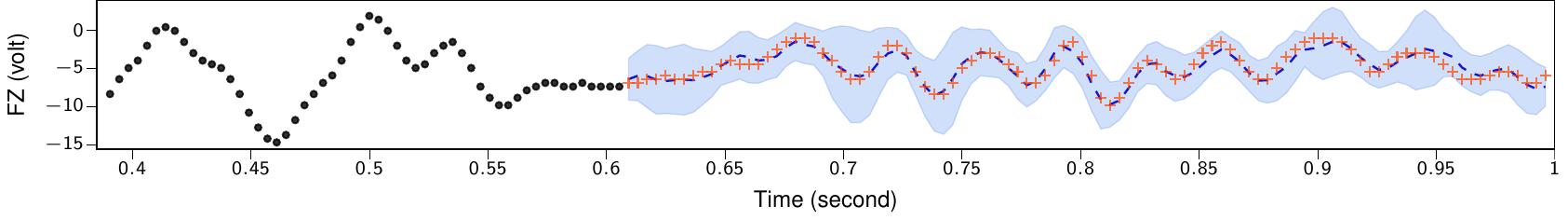}
        \tikzexternalenable

        \caption{\gls{SGPBAE}}
    \end{subfigure}

    \vspace{2ex}
    \centering
    {   
        \scriptsize
        \definecolor{color_observed}{HTML}{333333}
\definecolor{color_missing}{HTML}{fa5316}
\definecolor{color_mean}{HTML}{1416d1}
\definecolor{color_std}{HTML}{c8daf9}
\definecolor{color_std_2}{HTML}{b2cbf6}

\tikzexternaldisable
{\setlength{\tabcolsep}{2.8pt}
    \begin{tabular}{rlrlrlrl}
        \toprule
        {\protect\tikz[baseline=-1ex]\protect\draw[color=color_observed, mark=*, fill=color_observed, opacity=0.99, mark size=1.3pt, line width=2.1pt] plot[] (-0.1,0);} & \textsf{Observed values} &
        {\protect\tikz[baseline=-1ex]\protect\draw[color=color_missing, fill=color_missing, mark=+, opacity=0.99, mark size=2.2pt, line width=1.0pt] plot[] (-0.1,0);} & \textsf{Missing values} &
        {\protect\tikz[baseline=-1ex]\protect\draw[dashed, color=color_mean, fill=color_mean, opacity=0.99, mark size=1.7pt, line width=1.9pt] plot[] (-0.0,0)--(-0.60,0);} & \textsf{Predicted mean} &
        {\protect\tikz[baseline=-1ex]\protect\draw[color=color_std_2, fill=color_std, opacity=0.99, mark size=1.7pt, line width=8.1pt] plot[] (-0.0,0)--(-0.55,0);} & \textsf{ $\pm$ 3 standard deviation} \\
        \bottomrule
    \end{tabular}\tikzexternalenable
}
    }
    
    \caption{Visualization of predictions for missing data of the \textsc{eeg} dataset.
            Each panel shows one of the three channels with missing data (orange crosses) and observed data (black points).
            \label{fig:eeg_vis}
            }
\end{figure}

\cref{fig:eeg_vis} depicts the predictive mean and uncertainty estimation for the missing values of the \textsc{eeg} dataset.

\end{document}